\documentclass{article} %
\usepackage{arxiv,times}

\newcommand{\eg}{\textit{e.g.}}
\newcommand{\ie}{\textit{i.e.}}
\newcommand{\modelname}{\texttt{FlashWorld}}

\usepackage{amsmath,amsfonts,bm}

\def\eqref#1{equation~\ref{#1}}

\def\1{\bm{1}}

\DeclareMathAlphabet{\mathsfit}{\encodingdefault}{\sfdefault}{m}{sl}
\SetMathAlphabet{\mathsfit}{bold}{\encodingdefault}{\sfdefault}{bx}{n}

\usepackage{hyperref}
\definecolor{cvprblue}{rgb}{0.21,0.49,0.74}
\hypersetup{colorlinks,urlcolor=black,citecolor=cvprblue}
\usepackage{url}

\usepackage{booktabs}       %
\usepackage{amsfonts}       %
\usepackage{nicefrac}       %
\usepackage{microtype}      %
\usepackage{graphicx}
\usepackage{tabularx}
\usepackage{multicol}
\usepackage{multirow}
\usepackage{tabu}
\usepackage{amsmath}
\usepackage{xcolor,colortbl}
\usepackage{xspace}
\usepackage{subcaption}
\usepackage{sidecap}
\usepackage{afterpage}
\usepackage{floatpag}

\usepackage{makecell}

\title{FlashWorld: High-quality 3D Scene Generation within Seconds}

\author{Xinyang Li$^{1}$\thanks{Work done during an internship at Tencent.}~, Tengfei Wang$^{2}$, Zixiao Gu$^{3}$, Shengchuan Zhang$^{1}$, Chunchao Guo$^{2}$, Liujuan Cao$^{1}$\thanks{Corresponding Authors.} \\
$^1$ Key Laboratory of Multimedia Trusted Perception and Efficient Computing, Ministry of Education\\~~~of China, Xiamen University, $^2$ Tencent, $^3$ Yes Lab, Fudan University\\
\\
Project Page: \href{https://imlixinyang.github.io/FlashWorld-Project-Page/}{\texttt{\textcolor{blue}{https://imlixinyang.github.io/FlashWorld-Project-Page/}}}
}

\iclrfinalcopy %
\begin{document}

\maketitle

\begin{abstract}
We propose \modelname, a generative model that produces 3D scenes from a single image or text prompt in seconds, $10 \sim 100\times$ faster than previous works while possessing superior rendering quality.
Our approach shifts from the conventional multi-view-oriented \textit{(MV-oriented)} paradigm, which generates multi-view images for subsequent 3D reconstruction, to a \textit{3D-oriented} approach where the model directly produces 3D Gaussian representations during multi-view generation.
While ensuring 3D consistency, 3D-oriented method typically suffers poor visual quality.
\modelname~includes a dual-mode pre-training phase followed by a cross-mode post-training phase, effectively integrating the strengths of both paradigms.
Specifically, leveraging the prior from a video diffusion model, we first pre-train a dual-mode multi-view diffusion model, which jointly supports MV-oriented and 3D-oriented generation modes. 
To bridge the quality gap in 3D-oriented generation, we further propose a cross-mode post-training distillation by matching distribution from consistent 3D-oriented mode to high-quality MV-oriented mode. 
This not only enhances visual quality while maintaining 3D consistency, but also reduces the required denoising steps for inference.
Also, we propose a strategy to leverage massive single-view images and text prompts during this process to enhance the model's generalization to out-of-distribution inputs.
Extensive experiments demonstrate the superiority and efficiency of our method.
\end{abstract}

\begin{figure}[h]
    \centering
    \vspace{-12pt}
    \includegraphics[width=1\linewidth]{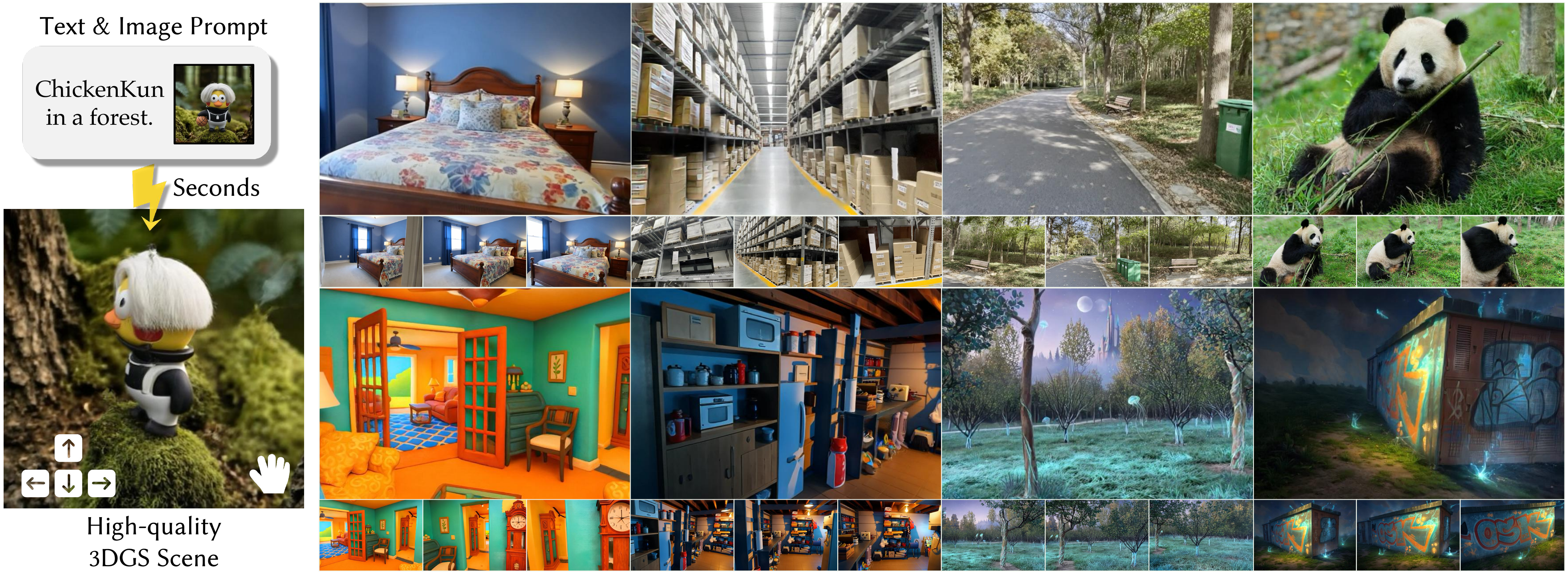}
    \caption{
    \modelname~enables fast and high-quality 3D scene generation across diverse scenes.
    }
    \label{fig:teaser}
\end{figure}

\section{Introduction}

3D generation shows great promise for applications in gaming, robotics, and VR/AR. However, generating full 3D scenes remains a significant challenge for both quality and efficiency, compared to generating individual 3D objects. These challenges stem from two core obstacles: the scarcity of high-quality 3D scene data and the exponential complexity of modeling real-world scenes.

\begin{figure}[t]
    \newcommand{\fourwide}{3.3cm}
    \begin{tabular}{c@{\,\,}c@{\,\,}c@{\,\,}c@{\,\,}}
    Input
    & CAT3D
    & Bolt3D
    & Wonderland
    \\
    \includegraphics[width=\fourwide]{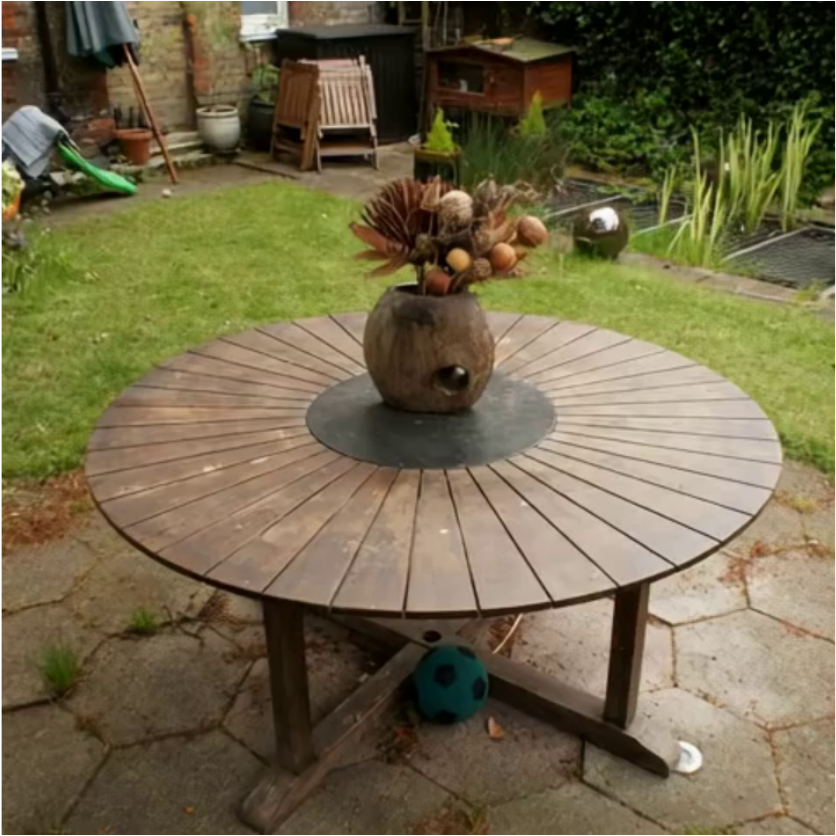}
    & \includegraphics[width=\fourwide]{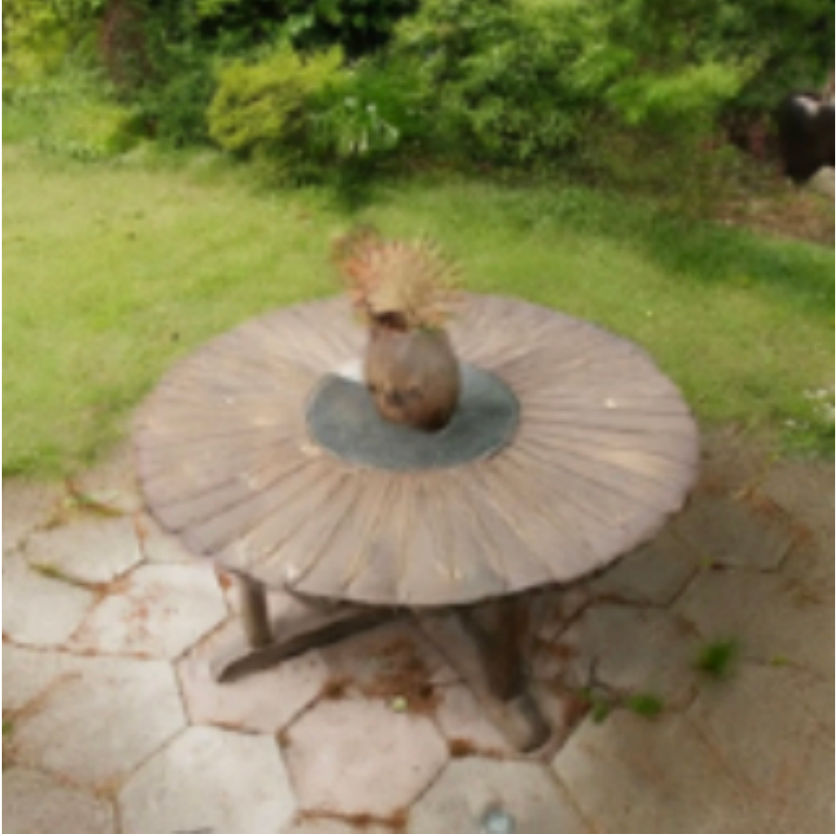}
    & \includegraphics[width=\fourwide]{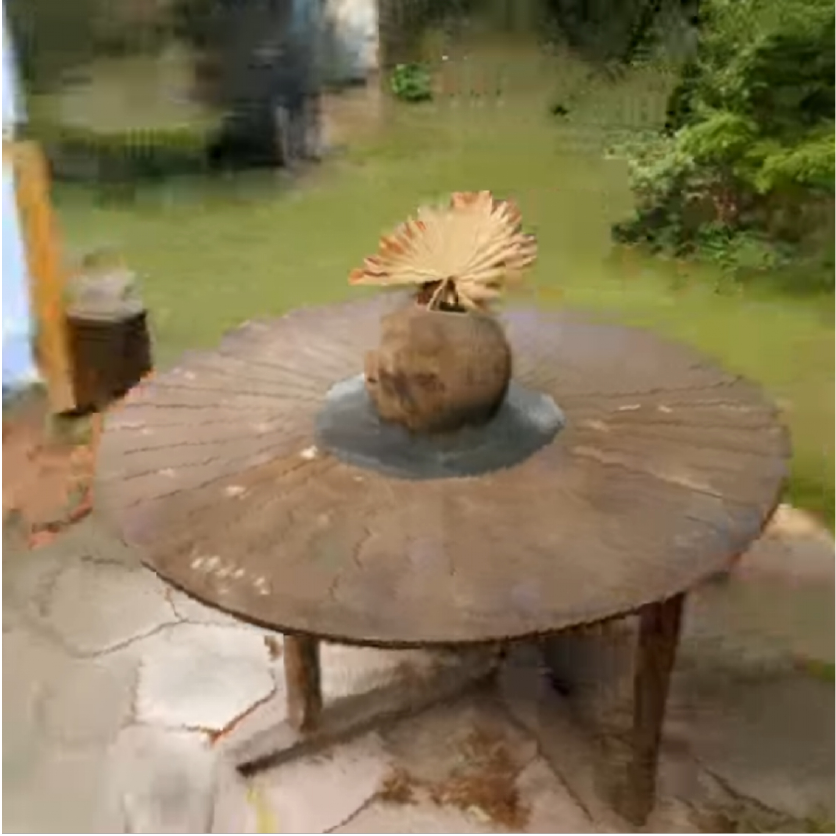}
    & \includegraphics[width=\fourwide]{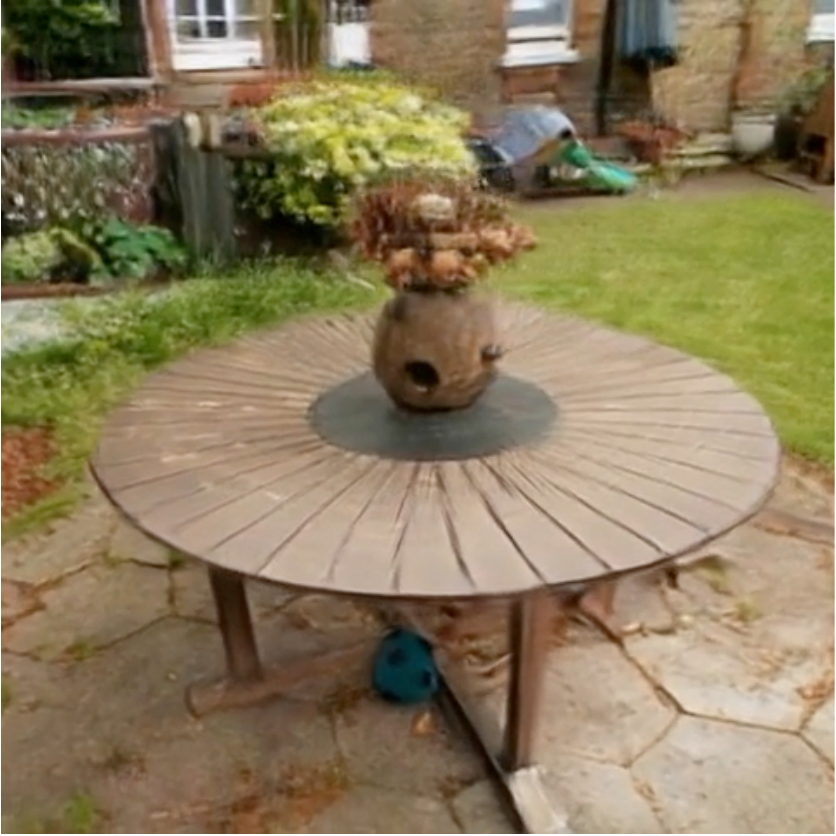}
    \\
    & 77 min (A100 GPU)
    & 15 sec (A100 GPU)
    & 5 min (A100 GPU)
    \\
    \vspace{-4mm}\\
    \cmidrule{1-4}
    \vspace{-3mm}\\
    Ours \textit{w/} MV-Diff
    & Ours \textit{w/} 3D-Diff
    & Ours \textit{w/} MV-Dist
    & Ours
    \\
    \includegraphics[width=\fourwide]{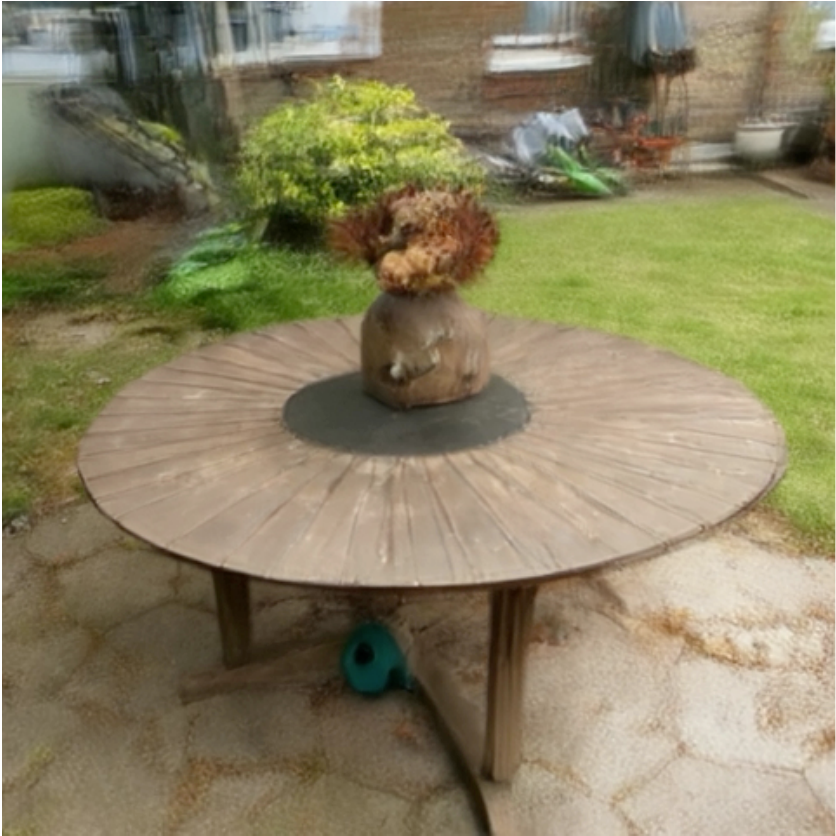}
    & \includegraphics[width=\fourwide]{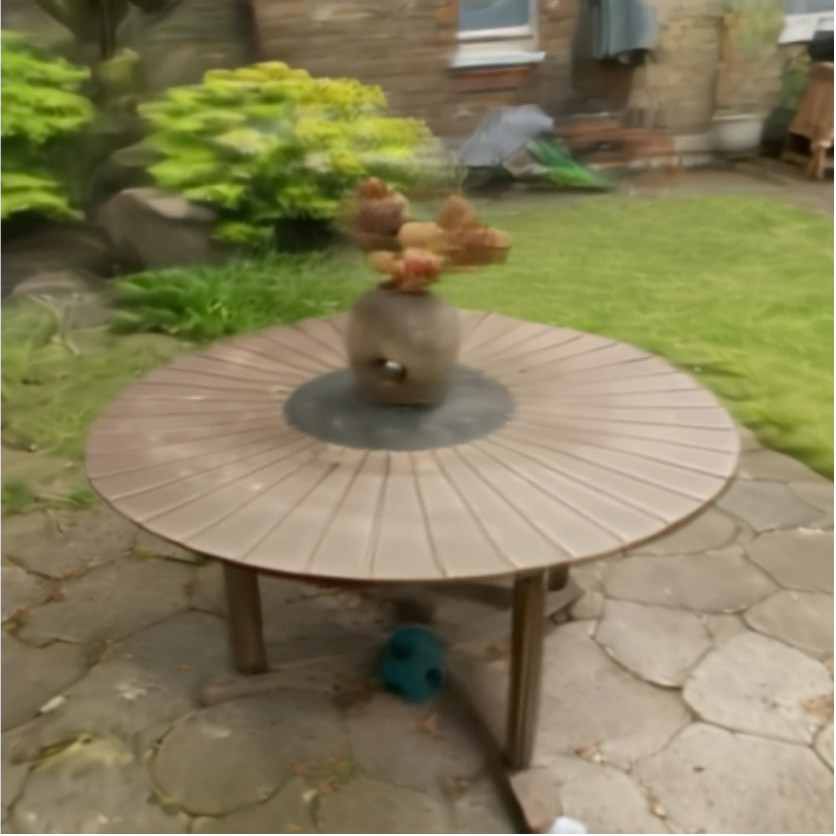}
    & \includegraphics[width=\fourwide]{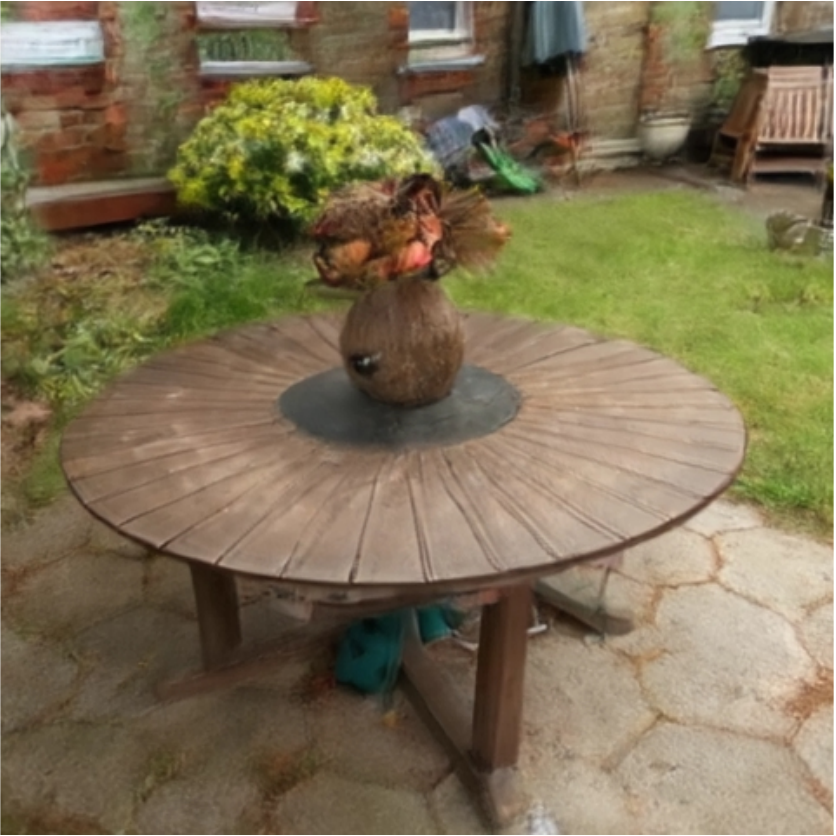}
    & \includegraphics[width=\fourwide]{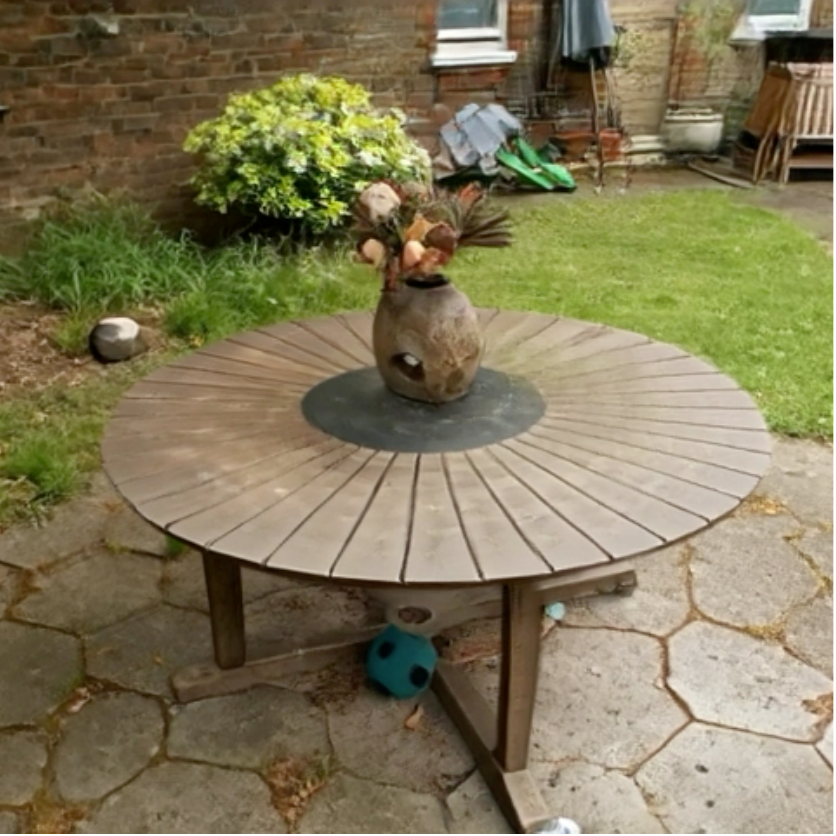}
    \\
    1 min (H20 GPU)
    & 1 min (H20 GPU)
    & 7 sec (H20 GPU)
    & 9 sec (H20 GPU)
    \\
    \end{tabular}
    \vspace{-3pt}
    \caption{{\bf A brief comparison of different 3D scene generation methods.} 
    MV-oriented diffusion methods (\ie, CAT3D~\citep{gao2024cat3d}, Bolt3D~\citep{szymanowicz2025bolt3d}, Wonderland~\citep{liang2025wonderland}, and Ours \textit{w/} MV-Diff) suffer from noisy textures due to multi-view inconsistency. 
    MV-oriented distillation further exacerbates this flaw (\ie, Ours \textit{w/} MV-Dist).
    3D-oriented diffusion methods (\ie, Ours \textit{w/} 3D-Diff) suffer from blurry visual effect.
    Our cross-mode distillation model (\ie, Ours) simultaneously solves these, making the quality of the novel view close to the input view.
    The time cost per scene, tested on a single GPU, is presented at the bottom of each method.
    }
    \label{fig:motivation}
\end{figure}

Early methods typically relied on assembling pre-existing 3D assets~\citep{xu2002constraint,yu2011make,wu2018miqp,feng2023layoutgpt,ccelen2024design,yang2024holodeck,deng2025global} or iteratively reconstructing scenes from inpainted images and depth maps~\citep{cai2023diffdreamer,fridman2023scenescape,hollein2023text2room,lei2023rgbd2,yu2024wonderjourney,zhang2024text2nerf,zhang20243d,chung2023luciddreamer,yu2025wonderworld,shriram2025realmdreamer,ni2025wonderturbo}.
Yet, without holistic scene-level understanding or multi-view consistency constraints, these approaches often struggle to produce semantically coherent and visually realistic scenes.
To address this, scalable data-driven approaches have emerged. The dominant paradigm is a two-stage, multi-view-oriented (MV-oriented) pipeline~\citep{gao2024cat3d,sun2024dimensionx,wallingford2024image,zhaogenxd,szymanowicz2025bolt3d,yang2025prometheus,go2025splatflow,go2025videorfsplat}: a diffusion model first generates multiple views from text or reference images, and then a 3D reconstruction is performed. 
However, the lack of explicit 3D constraints during view synthesis often causes geometric and semantic inconsistencies in generated views, leading to a noticeable visual quality gap between synthesized views and the reconstructed 3D scene.
Moreover, the considerable computational overhead of both the diffusion and reconstruction stages leads to generation latencies of several minutes to hours, as shown in Fig.~\ref{fig:motivation}. 
These limitations compromise the effectiveness and efficiency of current 3D scene generation methods, blocking their applications.

One promising but relatively less explored direction is the 3D-oriented scene generation pipeline~\citep{xu2023dmv3d,li2024dual3d,li2024director3d,tang2025cycle3d,cai2024baking}. 
These methods combine differentiable rendering~\citep{mildenhall2021nerf,wang2021neus,kerbl3Dgaussians} with diffusion models, allowing for direct 3D scene generation without an additional reconstruction stage. 
However, these generated 3D scenes often suffer from visual artifacts and blurry content. 
Consequently, they often require an additional refinement stage, which significantly degrades generation efficiency.

To enhance efficiency of diffusion models, post-training distillation techniques, such as consistency model distillation~\citep{song2023consistency} and distribution matching distillation~\citep{yin2024one,yin2024improved,xie2024distillation}, are often used. 
However, directly applying distillation amplifies each framework’s inherent limitations: \eg, it exacerbates multi-view inconsistency in the MV-oriented pipeline. 

In this work, we introduce a novel framework that combines the strengths of both paradigms through distillation, achieving substantial gains in 3D consistency and visual fidelity while significantly accelerating inference speed. 
Our contributions are briefly summarized as follows:

$\bullet$ We introduce a dual mode pretraining strategy built on a video diffusion model to train a multi-view diffusion model capable of operating in both MV-oriented and 3D-oriented modes. 

$\bullet$ We propose a cross-mode post-training strategy, where the MV-oriented mode serves as the teacher to improve visual quality, while the 3D-oriented mode acts as the student to ensure 3D consistency.

$\bullet$ To improve out-of-distribution generalization ability, we introduce a novel strategy that can leverage massive unlabeled image data and text prompts with randomly simulated camera trajectories during post-training, enhancing the model’s adaptability to diverse inputs, as shown in Fig.~\ref{fig:teaser}.

\section{Preliminary}
\noindent\textbf{Diffusion models}~\citep{ho2020denoising} generate data by progressively transforming samples from a standard Gaussian distribution $p(x_T) \sim \mathcal{N}(\mathbf{0}, \mathbf{I})$ into samples from a target data distribution $p(x)$, which have been widely applied across multiple domains, including image synthesis~\citep{rombach2022high}, multi-view generation~\citep{shi2023mvdream,tang2023mvdiffusion}, video generation~\citep{blattmann2023stable,wan2025wan}, and panoramic 3D scenes~\citep{hunyuanworld2025tencent}. 
The core methodology involves training a denoising network with optimizable parameters to reconstruct the original data by removing the injected Gaussian noise $\epsilon$ from $x$ according to a predefined noise schedule.
The forward process is formulated as:
$
x_t = F(x, t) = \alpha_t x + \sigma_t \epsilon,
$
where $\alpha_t$ and $\sigma_t$ jointly control the signal-to-noise ratio at each timestep $t$.
The denoising network can be trained to predict clean data $\hat{x}$ from noisy input $x_t$ by minimizing the following objective:
\begin{equation}
\mathcal{L} = \mathbb{E}_{x, t, \epsilon} \left[ \left\| x - \hat{x}_\theta(x_t, t) \right\|^2 \right].
\label{eq:diffusion_loss}
\end{equation}
Alternative training objectives include predicting noise $\epsilon$~\citep{ho2020denoising} or a linear combination of $x_0$ and $\epsilon$, known as $v$-prediction~\citep{salimans2022progressive}.
All predictions can be converted to the denoised estimate $\mu(x_{t}, t)$ and represent the gradient of the log probability of the distribution:
\begin{equation}
s(x_t, t)=\nabla_{x_t}\log p_{t}(x_t)=-\frac{x_t - \alpha_t \mu(x_t, t)}{\sigma_t^2}.
\label{eq:score_gradient}
\end{equation}

\noindent\textbf{Distribution matching distillation (DMD)}~\citep{yin2024one,yin2024improved} is an advanced technique designed to distill a slow, multi-step teacher diffusion model into a fast, few-step student model with comparable generation capabilities.
The key component is to minimize the approximate KL divergence across randomly sampled timesteps $t$ and noise inputs $z$ between the smoothed real data distribution $p_{\text{real}}(x_t)$ and the student generator’s output distribution $p_{\text{fake}}(x_t)$ by:
\begin{equation}
\nabla \mathcal{L}_{\text{DMD}} = -\mathbb{E}_t \left( \int \left( s_{\text{real}}(F(G_{\theta}(z), t), t) - s_{\text{fake}}(F(G_{\theta}(z), t), t) \right) \frac{dG_\theta(z)}{d\theta} dz \right),
\label{eq:dmd}
\end{equation}
where $ s_{\text{real}} $ and $ s_{\text{fake}} $ are approximated scores using diffusion models $ \mu_{\text{real}} $ and $ \mu_{\text{fake}} $ trained on their respective distributions (Eq.~\ref{eq:diffusion_loss}). 
DMD uses a frozen pre-trained diffusion model $ \mu_{\text{real}} $ as the teacher, and dynamically updates $ \mu_{\text{fake}} $ while training $G_\theta$, using diffusion loss on samples from the generator.

\section{Method} 

The core of our framework lies in leveraging DMD to transfer knowledge from a MV-oriented multi-view diffusion model, one well-established for high visual quality, to a 3D-oriented few-step multi-view generator, which is inherently endowed with 3D consistency.
However, this paradigm introduces two key challenges:
First, for open-world 3D scene generation, the 3D-oriented few-step generator requires a sufficiently robust prior and strong generative capacity from the start. Without this, the training process is prone to collapse.
Second, due to the limited quantity and diversity of high-quality multi-view datasets, it becomes critical to develop a strategy that effectively handles scenarios with diverse styles, object categories, and camera trajectories.
Specifically, To address these challenges, we first design a dual-mode pre-training strategy as detailed in Sec.~\ref{sec:dual-mode}. This strategy yields a multi-view diffusion model that operates in two distinct modes: a MV-oriented mode for high visual fidelity and a 3D-oriented mode for inherent 3D consistency. 
Subsequently, in Sec.~\ref{sec:cross-mode}, we present a cross-mode post-training framework to bridge these two modes: the MV-oriented mode acts as the \textit{teacher}, supplying score distillation gradients to ensure visual quality; the 3D-oriented mode serves as the \textit{student}, learning to inherit the teacher's distribution while preserving 3D consistency.
Furthermore, to explicitly tackle out-of-distribution generalization, in Sec.~\ref{sec:ood}, we introduce a strategy that can leverage single-view image data, text prompts, and pre-defined camera trajectories, boosting the model’s adaptability to diverse scenarios.

\begin{figure}[t]
    \centering
    \includegraphics[width=1\linewidth]{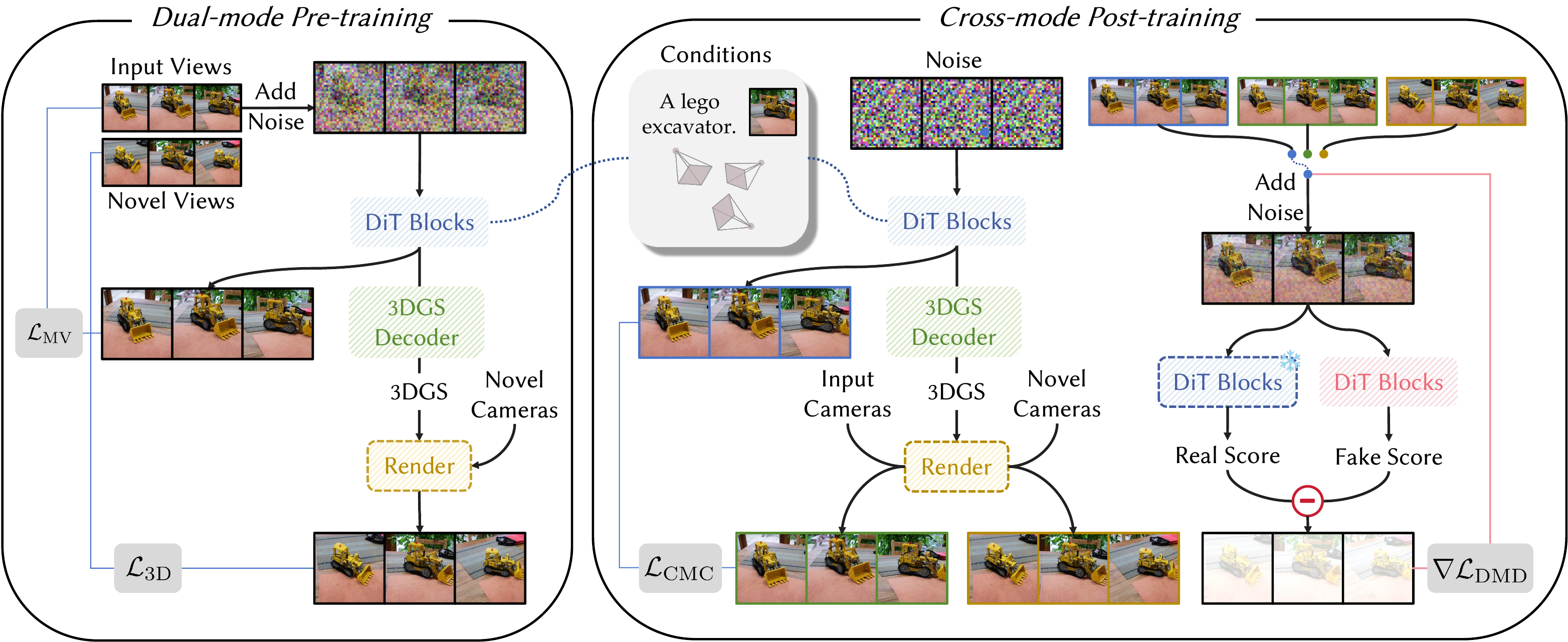}
    \caption{{\bf Method overview.} We first pre-train a dual-mode multi-view latent diffusion model using multi-view datasets, and then employ an cross-mode distillation post-training strategy to accelerate generation while enhancing visual quality and inheriting 3D consistency.}
    \label{fig:method}
\end{figure}

\subsection{Dual-mode Pre-training}
\label{sec:dual-mode}

In this stage, we pre-train a dual-mode multi-view latent diffusion model using multi-view datasets, as illustrated in Fig.~\ref{fig:method} (left). For each training iteration, we sample a batch containing multi-view images $\mathcal{X}$, their corresponding camera parameters $\mathcal{C}$, and additional conditioning information $y$ (such as a text prompt or a single-view image). The multi-view images are first encoded into the latent space to obtain multi-view latents $\mathcal{Z} = E(\mathcal{X})$. A forward diffusion process is then applied to produce noisy multi-view latents $\mathcal{Z}_t = \alpha_t \mathcal{Z} + \sigma_t \epsilon$ at a randomly sampled timestep $t$.

The noisy latents $\mathcal{Z}_t$, together with the camera parameters $\mathcal{C}$ and conditioning $y$, are input to the denoising network for reverse denoising training. We represent cameras using Reference-Point Plücker Coordinates~\citep{cai2024baking} raymaps. The denoising network is a Diffusion Transformer (DiT)~\citep{peebles2023scalable} enhanced with 3D attention blocks, and outputs both a denoised estimate $\hat{\mathcal{Z}}_{\text{MV}}$ and an auxiliary multi-view feature $\mathcal{F}$.
For the MV-oriented mode, we optimize:
\begin{equation}
\mathcal{L}_{\text{MV}} = \mathbb{E}_{\mathcal{X}, t, \epsilon, y, \mathcal{C}} \left[ \left\| \mathcal{Z} - \hat{\mathcal{Z}}_{\text{MV}} \right\|^2 \right].
\end{equation}
To enable 3D-oriented generation, we decode 3D Gaussian parameters from the multi-view feature $\mathcal{F}$ using a 3DGS decoder:
$
\{\tau, \boldsymbol{q}, \boldsymbol{s}, \alpha, \boldsymbol{c}\} = D_\mathcal{G}(\mathcal{F}),
$
where $\tau$, $\boldsymbol{q}$, $\boldsymbol{s}$, $\alpha$, and $\boldsymbol{c}$ represent the depth, rotation quaternion, scale, opacity, and spherical harmonics coefficients of the 3D Gaussians, respectively.
The 3DGS decoder $D_\mathcal{G}$ is initialized from the original latent decoder $D$, with its first and last convolutional layers re-initialized to accommodate the additional features and output channels required for the Gaussian parameters. The predicted depth is then converted to pixel-aligned Gaussian points via $\boldsymbol{\mu} = \boldsymbol{o} + \tau \boldsymbol{d}$, where $\boldsymbol{o}$ and $\boldsymbol{d}$ denote the camera origin and ray direction, respectively.
For the 3D-oriented mode, we optimize the following loss:
\begin{equation}
    \mathcal{L}_{\text{3D}} = \mathbb{E}_{\mathcal{X}, t, \epsilon, y, \mathcal{C}} \left[ \left\| \mathcal{X}_{\text{novel}} - R(\mathcal{G}, \mathcal{C}_{\text{novel}}) \right\|^2 \right],
    \label{eq:3d_loss}
\end{equation}
where $R$ denotes the rendering operation, $\mathcal{G} = \{\boldsymbol{\mu}, \boldsymbol{q}, \boldsymbol{s}, \alpha, \boldsymbol{c}\}$ is the set of 3D Gaussians, and $\mathcal{X}_{\text{novel}}$, $\mathcal{C}_{\text{novel}}$ are the ground-truth novel-view images and their associated cameras, respectively.
During inference, both MV-oriented and 3D-oriented modes can be used for denoising~\citep{li2024dual3d,li2024director3d}. In particular, for the 3D-oriented mode, the model predicts the estimated clean multi-view latents as $\hat{\mathcal{Z}}_{\text{3D}} = E(R(\mathcal{G}, \mathcal{C}))$.

In contrast to previous methods~\citep{li2024dual3d,li2024director3d} that are initialized from image diffusion models~\citep{rombach2022high}, we initialize our framework with a video diffusion model~\citep{wan2025wan}. We observe that this video model not only converges more rapidly, but also features a powerful VAE with a higher compression rate, enabling support for a larger number of views (\ie, 24) and higher output resolutions (\ie, 480P).

\subsection{Cross-mode Post-training}
\label{sec:cross-mode}

After pre-training, we employ an asymmetric distillation strategy to accelerate generation while enhancing visual quality and inheriting 3D consistency, as shown in Fig~\ref{fig:method} (right).
Specifically, we observe that while the MV-oriented mode exhibits poor consistency, it can generate multi-view images with high visual quality; thus, we leverage the MV-oriented mode of our dual-mode multi-view latent diffusion model as the real teacher $\mu_{\text{real}}$: this teacher model is frozen, tasked with computing the real score gradient.
Another copy of the model $\mu_{\text{fake}}$ is dynamically updated to estimate the fake score corresponding to the current distribution of the distilled generator. 
Meanwhile, our few-step student model is initialized with the 3D-oriented mode of our dual-mode multi-view latent diffusion model.

Specifically, to generate a 3D scene, the 3D-oriented multi-view generation process alternates between denoising and noise injection steps to enhance sample quality~\citep{luo2023latent}. 
We first define a schedule of $N$ timesteps, denoted as $\{t_1, t_2, \cdots, t_N\}$, where $N$ is typically small (\eg, 4).
Starting from a randomly sampled noise $\mathcal{Z}_{t_1} = z  \sim \mathcal{N}(\mathbf{0}, \mathbf{I})$, we alternate between 3D-oriented denoising updates $\hat{\mathcal{Z}}_{t_i} = E(R(G_{\theta,\text{3D}}(\mathcal{Z}_{t_{i}}, t_i, y, \mathcal{C}), \mathcal{C}))$ and forward diffusion steps $\mathcal{Z}_{t_{i + 1}} = \alpha_{t_{i + 1}}\hat{\mathcal{Z}}_{t_i} + \sigma_{t_{i + 1}}\epsilon$ where $G_{\theta,\text{3D}}$ is the 3DGS generator and $\epsilon \sim \mathcal{N}(\mathbf{0}, \mathbf{I})$, until obtaining the 3D Gaussians at the final step (\ie, $G_{\theta,\text{3D}}(\mathcal{Z}_{t_{N}}, t_{N}, y, \mathcal{C})$). 
At each step, the multi-view denoising update is performed based on rendering, thereby ensuring that 3D consistency is maintained throughout the process.

During distillation training, we adopt the DMD2 algorithm~\citep{yin2024improved}, which includes a DMD objective (\ie, Eq.~\ref{eq:dmd}) and a standard non-saturating GAN objective~\citep{goodfellow2020generative},
where the logits value required by the GAN loss is obtained by adding an extra classification branch with several convolutional layers at the end of the fake score network.
We adopt the estimated R1 regularization~\citep{lin2025diffusion} to stabilize the GAN training.
The DMD objective and the GAN objective are employed to optimize both the input and novel views.

We also observe that relying solely on the above strategy can lead to the generation of scenes with unstable floating artifacts. We hypothesize that this instability arises from the challenges in optimizing with noisy gradients introduced by Gaussian rendering and latent encoding. 
To address this, during post-training, we additionally update an MV-oriented student model at a lower frequency. This model shares the same DiT backbone as the 3D-oriented student model. To encourage alignment between the two modes, we introduce a cross-mode consistency loss:
\begin{equation}
    \mathcal{L}_{\text{CMC}} = \mathbb{E}_{z, t, \epsilon, y, \mathcal{C},i} \left[ \left\| E(R(G_{\theta,\text{3D}}(\mathcal{Z}_{t_{i}}, t_i, y, \mathcal{C}), \mathcal{C})) - G_{\theta,\text{MV}}(\mathcal{Z}_{t_{i}}, t_i, y, \mathcal{C}) \right\|^2 \right],
\end{equation}
where $\lambda$ is a small weighting factor (\ie, 0.1).
Because the MV-oriented mode prediction are less affected by unstable rendering gradients, this consistency loss regularizes the 3D-oriented mode to produce more stable and reliable generations.

\subsection{Out-of-Distribution Data Co-training. }
\label{sec:ood}

During pre-training, it is common to jointly train on image and video generation tasks to enhance the model's generalization ability. 
While this approach benefits the DiT backbone, it does not optimize the 3DGS decoder, potentially limiting the range of inputs the 3DGS decoder can effectively process.
To address this, in the post-training phase, we introduce a strategy to broaden the model's input distribution and improve generalization to diverse scenes, even when multi-view data is limited in quantity and variety. Specifically, we combine image or text conditions sampled from image datasets with random camera trajectories, which can be drawn either from multi-view sequences or from a set of predefined trajectories.
Importantly, we omit the GAN loss during this co-training process to prevent distribution mismatches.
This approach not only enhances the model's generalization to a wide range of input images and text prompts, but also increases its robustness when encountering out-of-distribution camera trajectories.
The details of this strategy are provided in Appendix~\ref{app:training_details} .

\section{Experiments}

In this section, we evaluate the performance of our method on various benchmarks, including image-to-3D scene generation, text-to-3D scene generation, and WorldScore benchmark.
For implementation details, please refer to Appendix~\ref{app:training_details}.

\subsection{Comparison on Image-to-3D Scene Generation}

\begin{figure}[t]
    \newcommand{\sidewide}{4.75cm}
    \newcommand{\midwide}{3.24cm}
    
    \begin{tabular}{c@{\,\,}c@{\,\,}c@{\,\,}c@{\,\,}}
    \rotatebox{90}{\quad\quad Ours} 
    & \includegraphics[width=\sidewide]{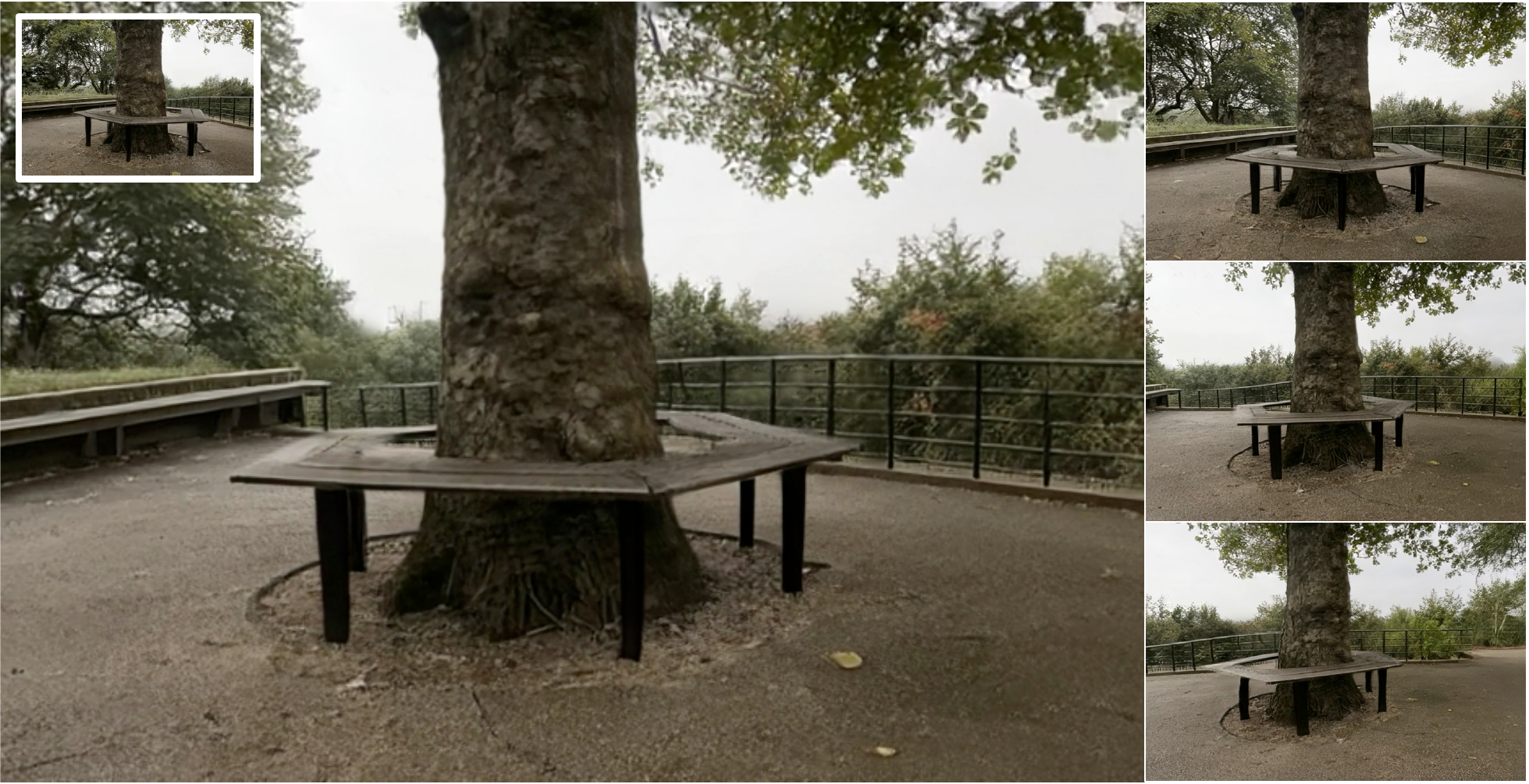} 
    & \includegraphics[width=\midwide]{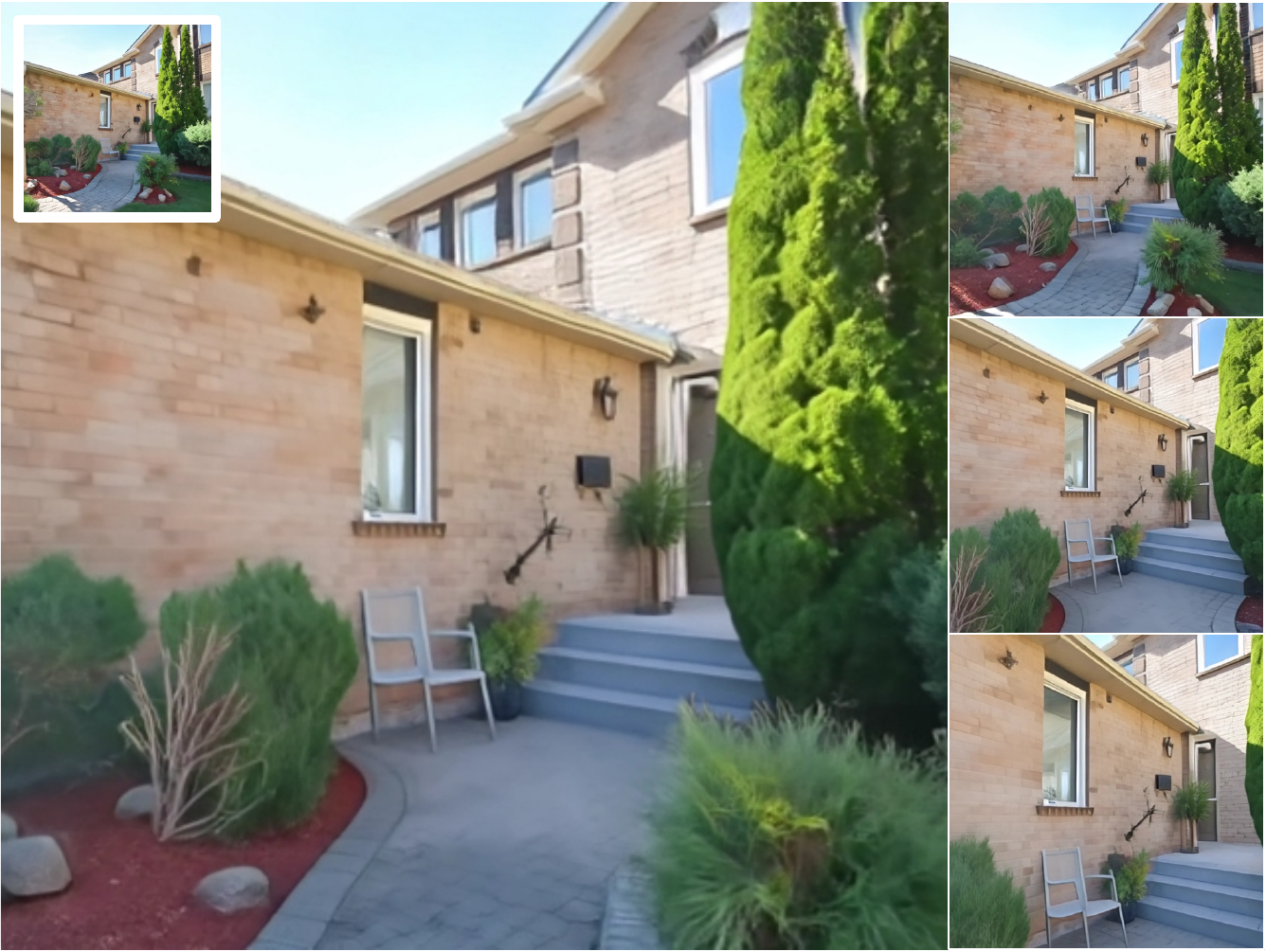}
    & \includegraphics[width=\sidewide]{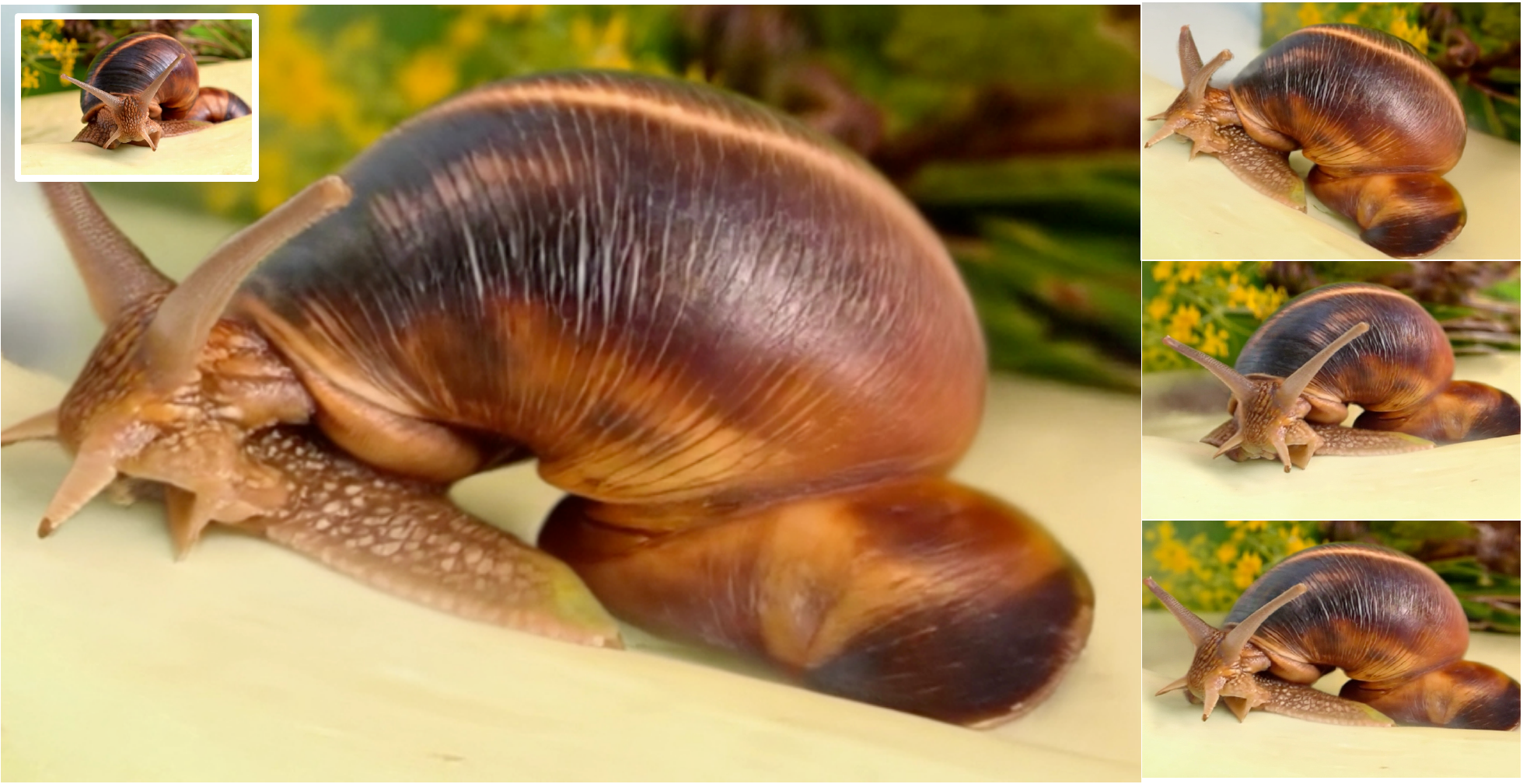}
    \\
    \vspace{-4mm}\\
    \cmidrule{1-4}
    \vspace{-3mm}\\
    & CAT3D
    & Bolt3D
    & Wonderland
    \\
    \vspace{-4mm}\\
    \rotatebox{90}{\quad Baselines} 
    & \includegraphics[width=\sidewide]{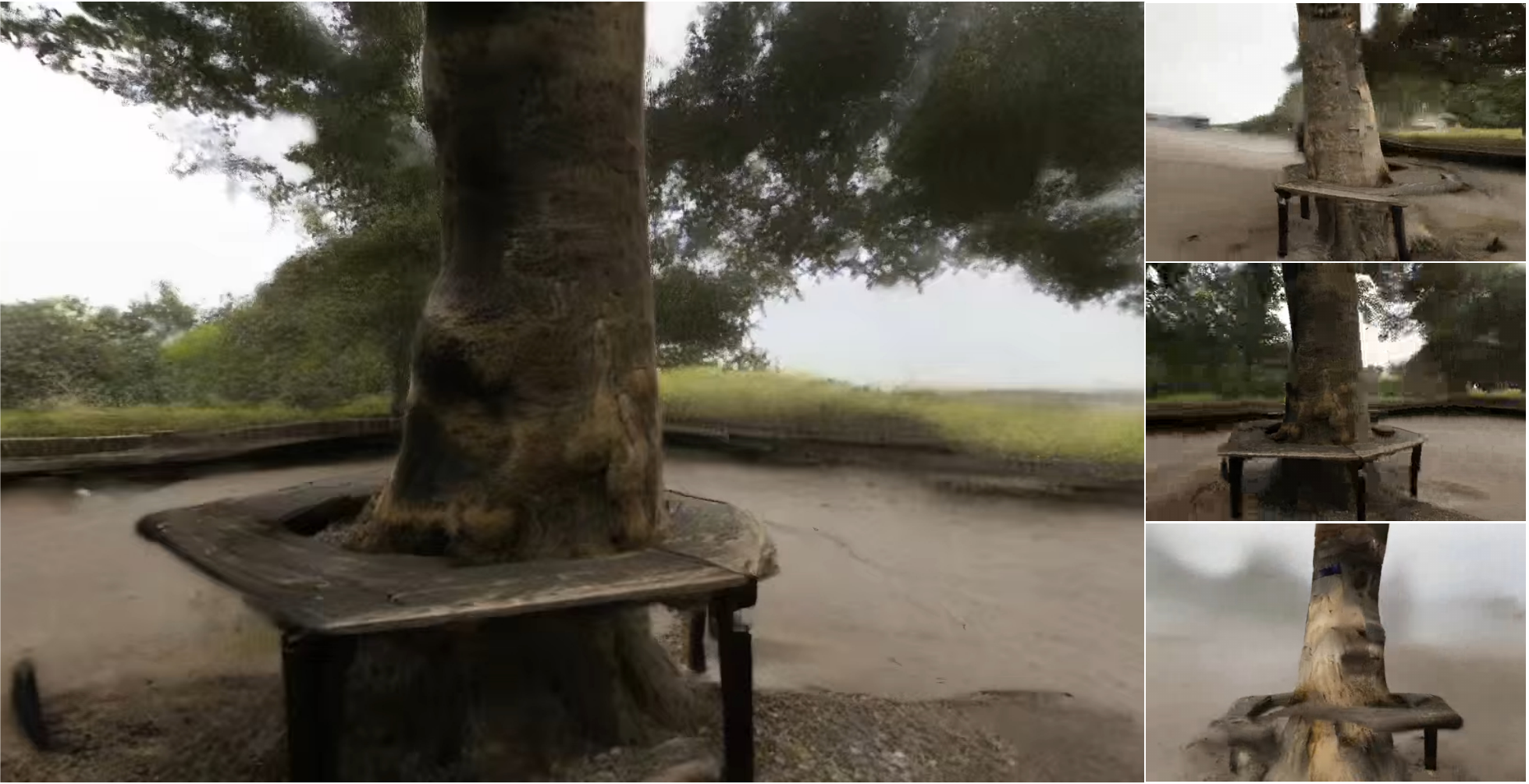} 
    & \includegraphics[width=\midwide]{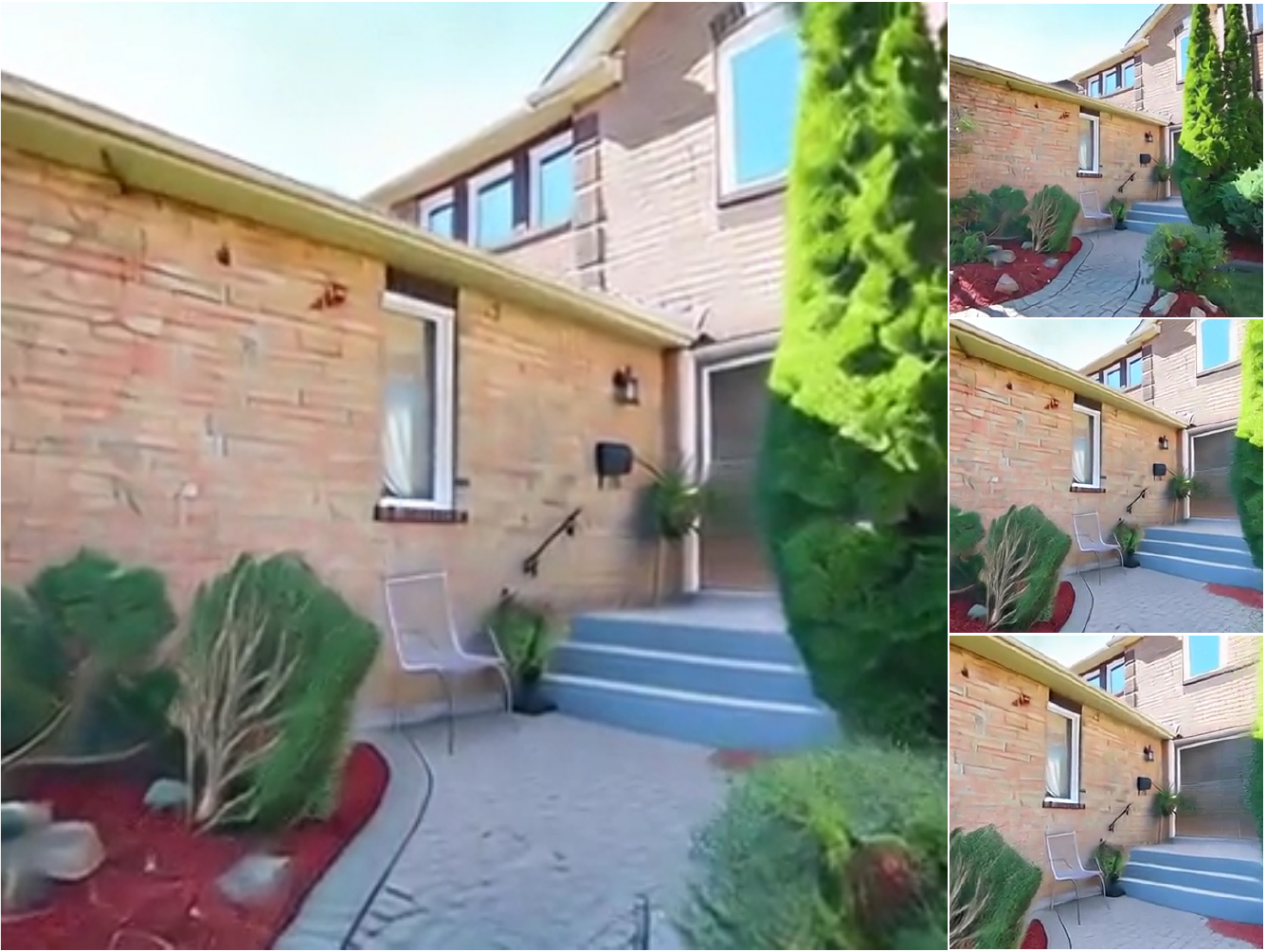}
    & \includegraphics[width=\sidewide]{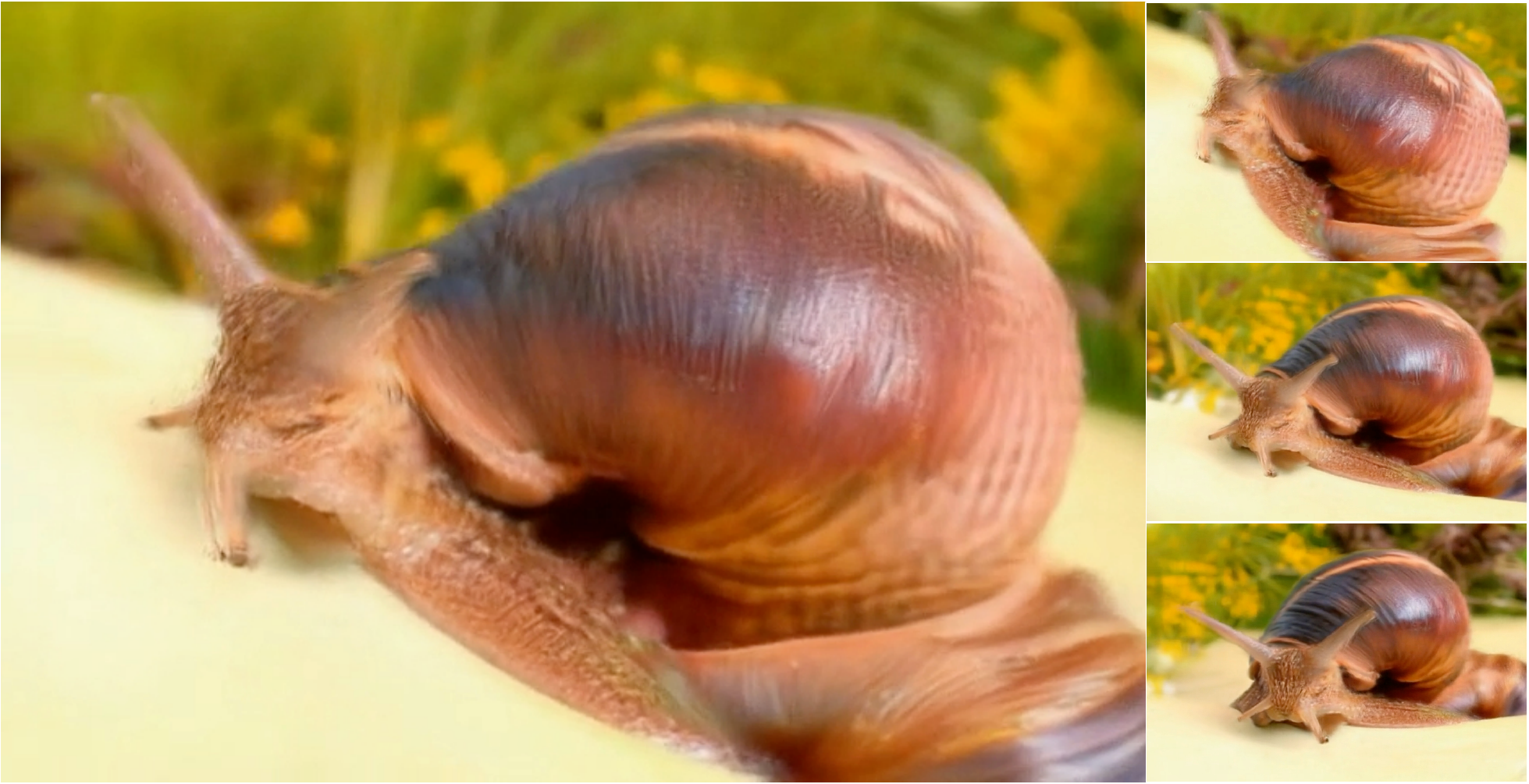} 
    \\
    \end{tabular}
    \caption{{\bf Image-to-3D scene generation results of different methods}.}
    \label{fig:qualitative_i23d}
\end{figure}

We present a qualitative comparison with state-of-the-art image-to-3D scene generation methods in Fig.~\ref{fig:qualitative_i23d}. 
These baselines are MV-oriented, including: CAT3D~\citep{gao2024cat3d}, which generates novel views via multi-view diffusion followed by optimization-based 3D reconstruction; Bolt3D~\citep{szymanowicz2025bolt3d}, which synthesizes both appearance and geometry for novel views and then applies a feed-forward 3D reconstruction; and Wonderland~\citep{liang2025wonderland}, a leading approach that leverages a powerful video diffusion model and latent-based feed-forward 3D reconstruction.
As these methods are not open-sourced, we utilize the video results provided in their respective project pages for visualization.
We employ ViPE~\citep{huang2025vipe} to estimate camera poses and intrinsics from the baseline videos.
CAT3D struggles to generate complex scenes, resulting in blurry outputs and missing geometric details. 
Bolt3D also exhibits inaccurate geometric details, such as imprecise tree branches and needle-like leaves.
Wonderland suffers from repeated and distorted Gaussian artifacts, especially under large camera pose changes.
Overall, these MV-oriented methods fail to generate complex scenes, primarily due to insufficient multi-view consistency.
In contrast, our model produces high-fidelity, detailed scenes and successfully recovers intricate structures (\eg, leaves, iron fences, and tentacles), highlighting the advantages of our 3D-oriented pipeline.

\subsection{Comparison on Text-to-3D Scene Generation}

\begin{figure}[t]
    \newcommand{\promptheight}{2.4cm}
    \newcommand{\fourwide}{3.1cm}
    
    \begin{tabular}{c@{\,\,}c@{\,\,}c@{\,\,}c@{\,\,}c@{\,\,}c@{\,\,}}
    \rotatebox{90}{Input} 
    & {\parbox[c][\promptheight][c]{\fourwide}{\centering \scriptsize\fontfamily{ppl}\selectfont\begin{tabular}{c}A fluffy,\\orange cat.
    \end{tabular}}}
    & {\parbox[c][\promptheight][c]{\fourwide}{\centering \scriptsize\fontfamily{ppl}\selectfont\begin{tabular}{c}A spacious kitchen\\ with wooden floors,\\ white countertops, and\\an island in the center.\end{tabular}}}
    & {\parbox[c][\promptheight][c]{\fourwide}{\centering \scriptsize\fontfamily{ppl}\selectfont\begin{tabular}{c}A traditional\\wooden gate with\\red lanterns.\end{tabular}}}
    & {\parbox[c][\promptheight][c]{\fourwide}{\centering \scriptsize\fontfamily{ppl}\selectfont\begin{tabular}{c}A large stone\\fountain surrounded\\by lush greenery\\and a clear blue sky.\end{tabular}}}
    \\
    \vspace{-10mm}\\
    \rotatebox{90}{\quad\quad\quad\quad\quad  Ours} 
    & \includegraphics[width=\fourwide]{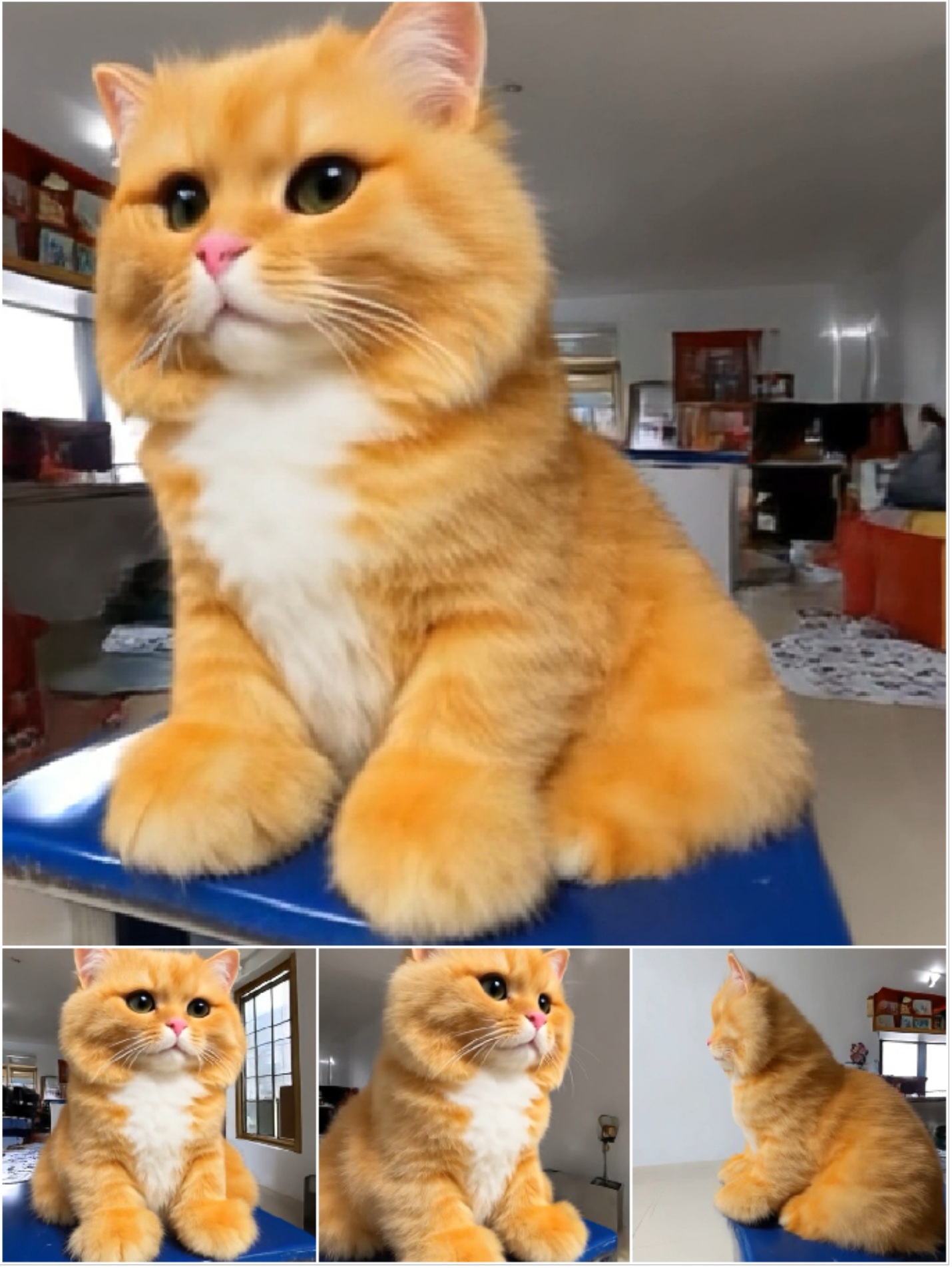}
    & \includegraphics[width=\fourwide]{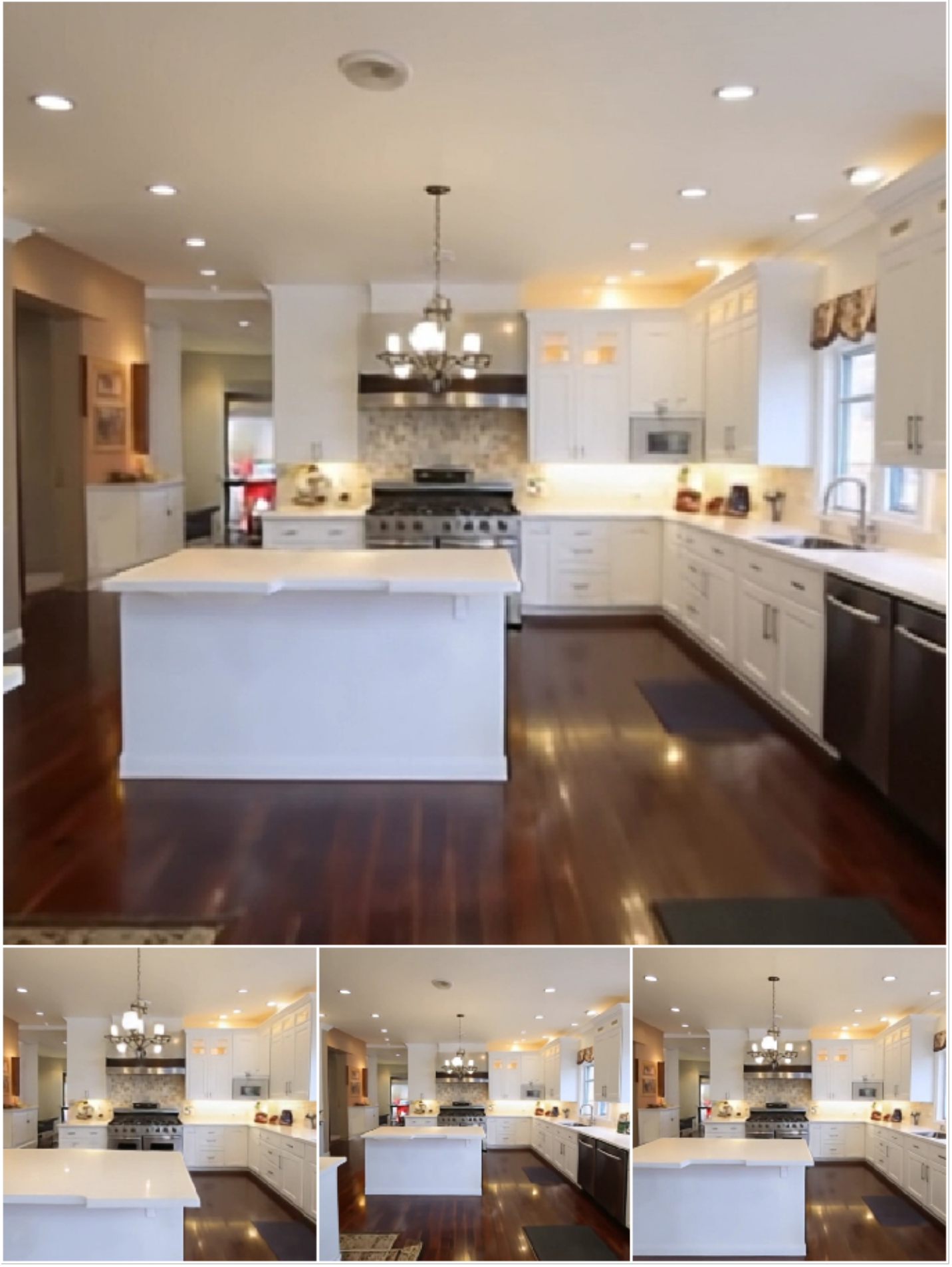}
    & \includegraphics[width=\fourwide]{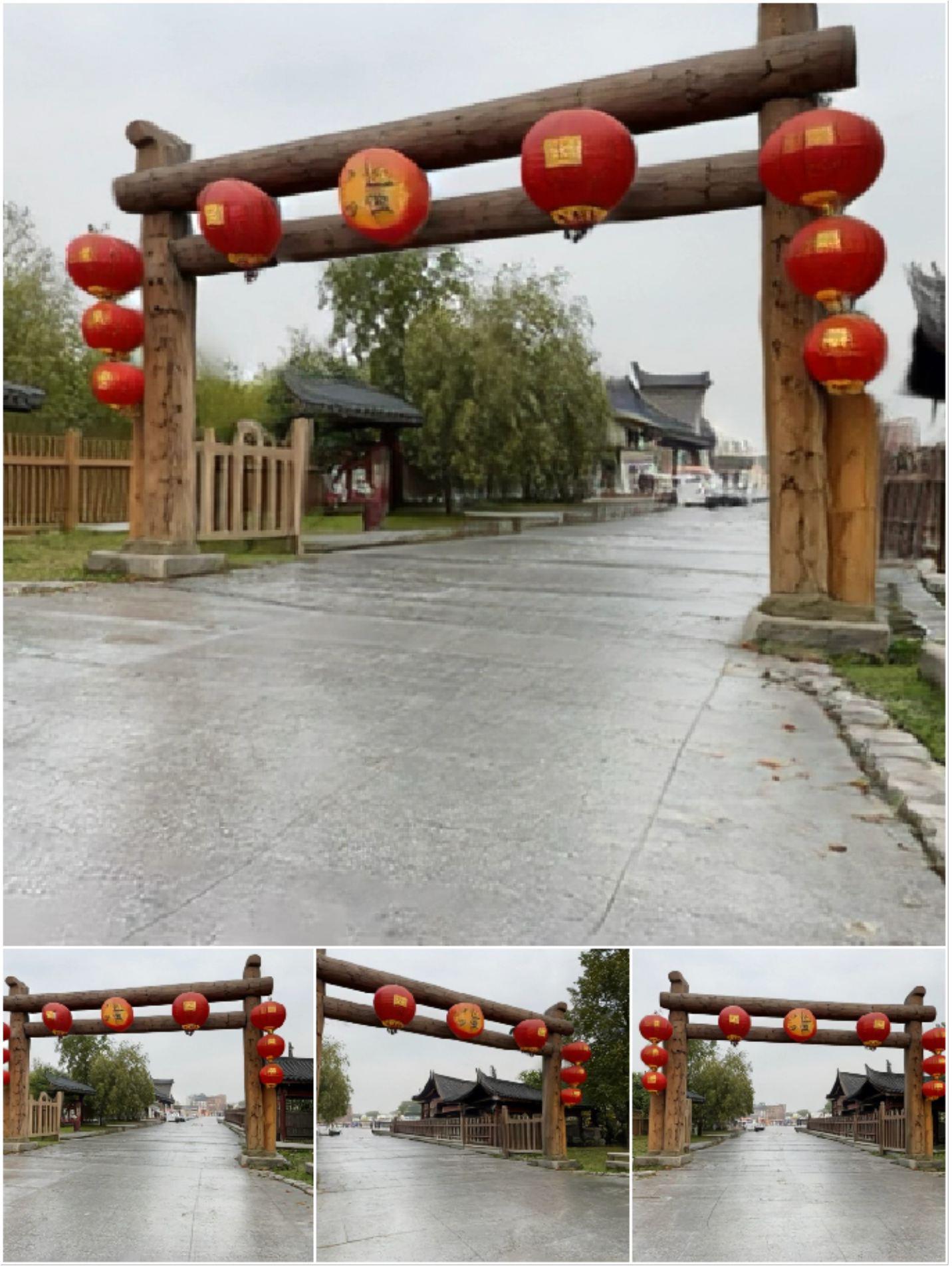}
    & \includegraphics[width=\fourwide]{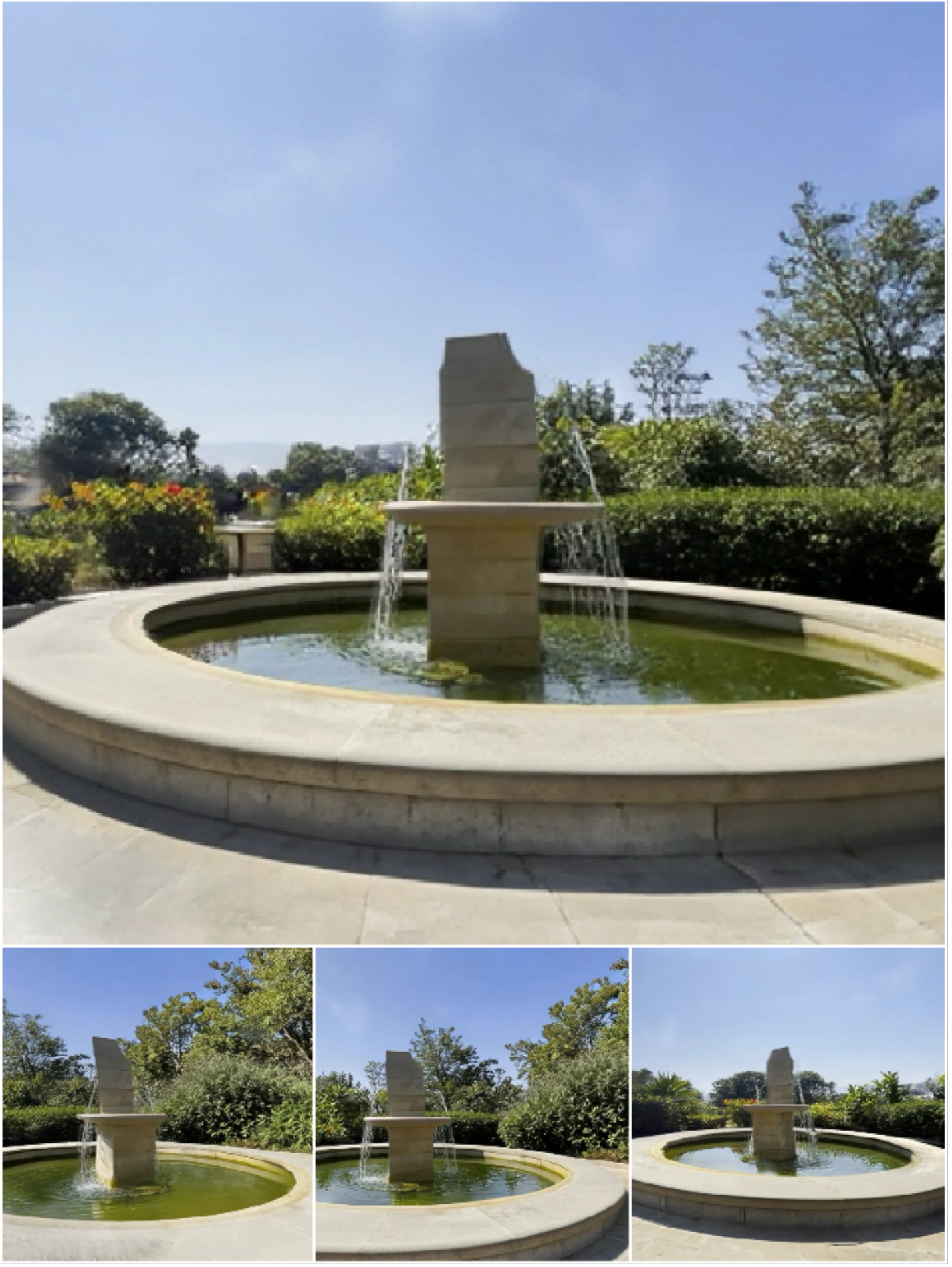}
    \\
    \vspace{-4mm}\\
    \cmidrule{1-5}
    \vspace{-3mm}\\
    & Director3D
    & Prometheus
    & SplatFlow
    & VideoRFSplat
    \\
    \rotatebox{90}{\quad\quad\quad\quad Baselines} 
    & \includegraphics[width=\fourwide]{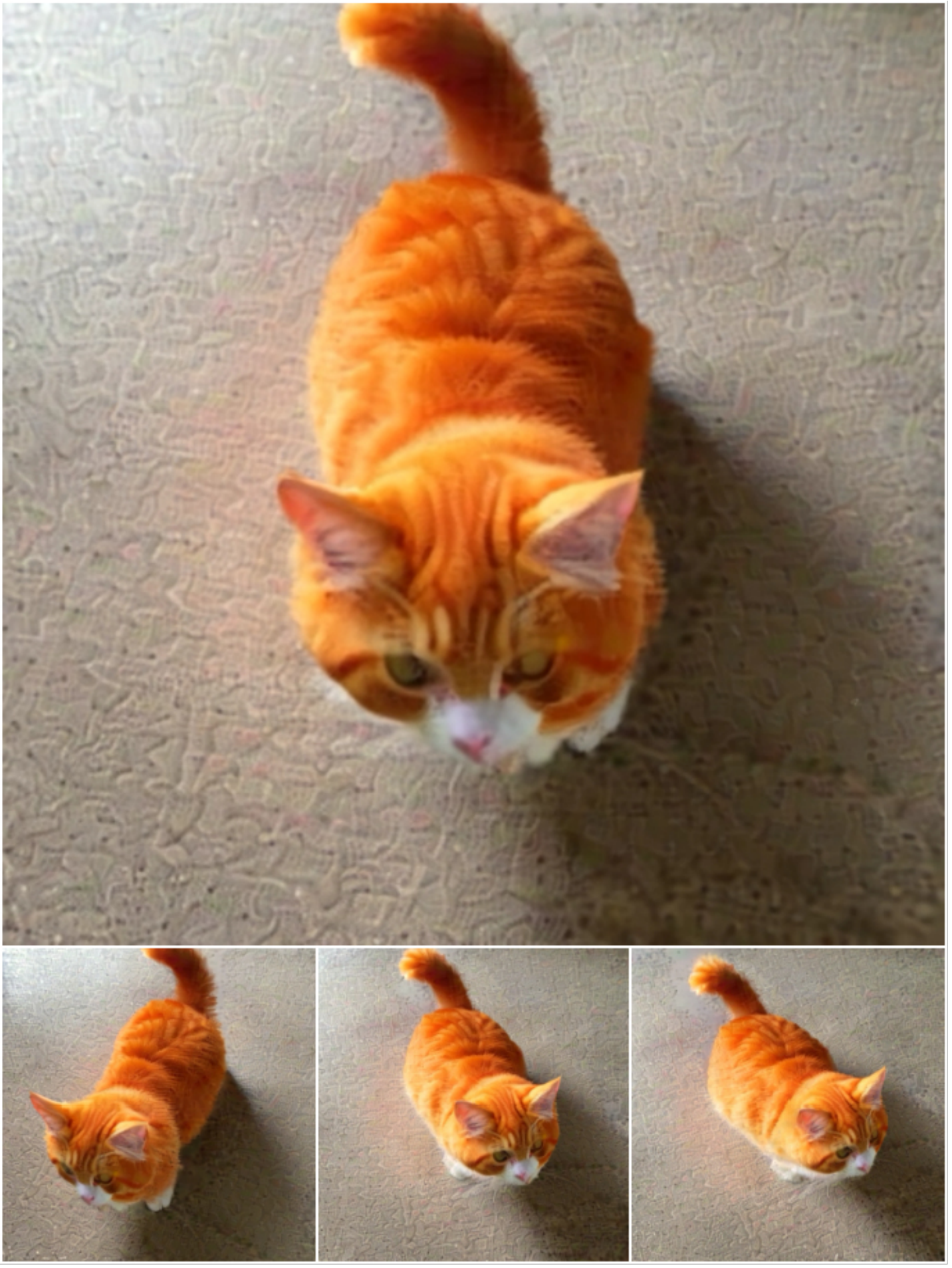}
    & \includegraphics[width=\fourwide]{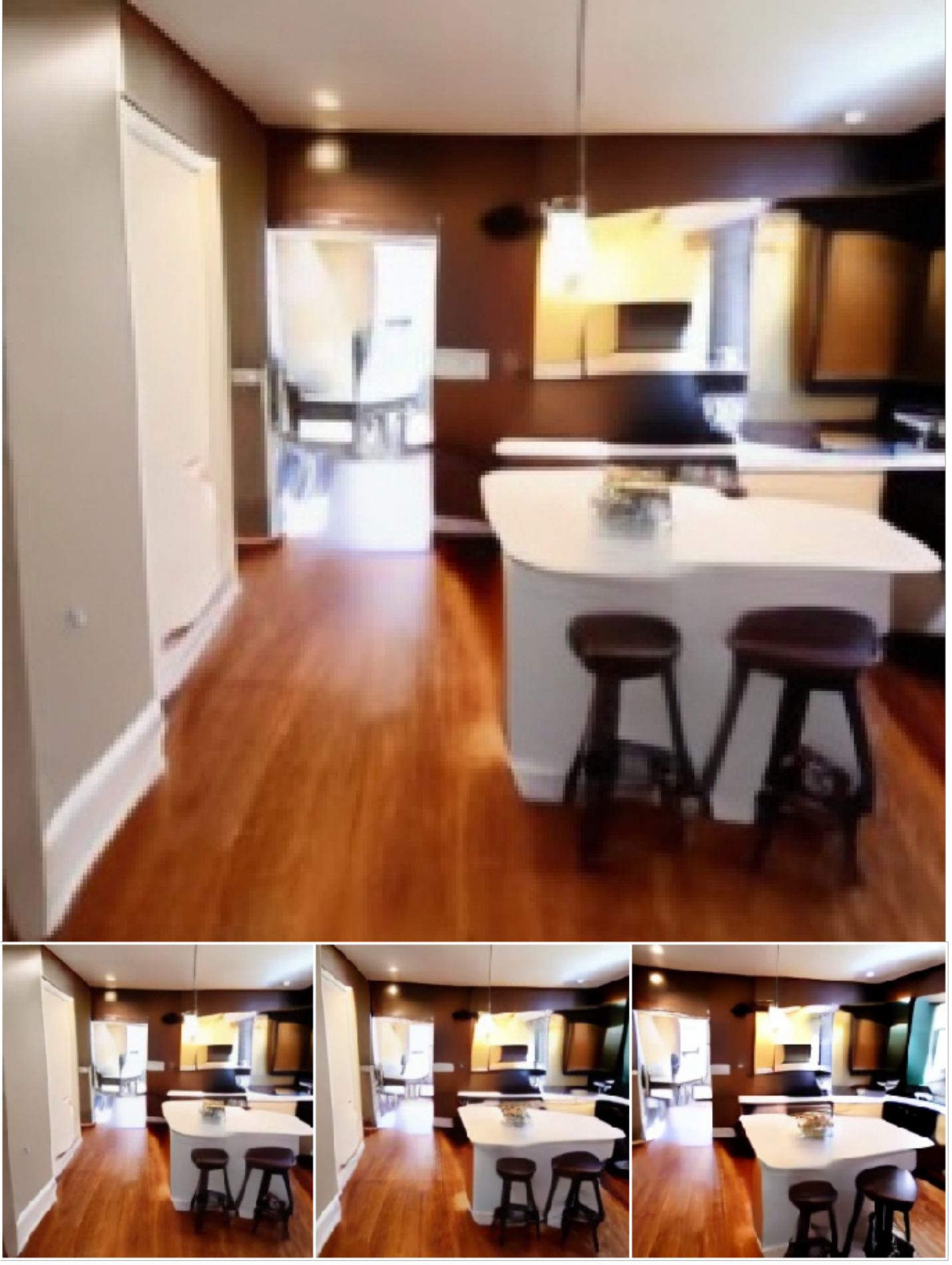}
    & \includegraphics[width=\fourwide]{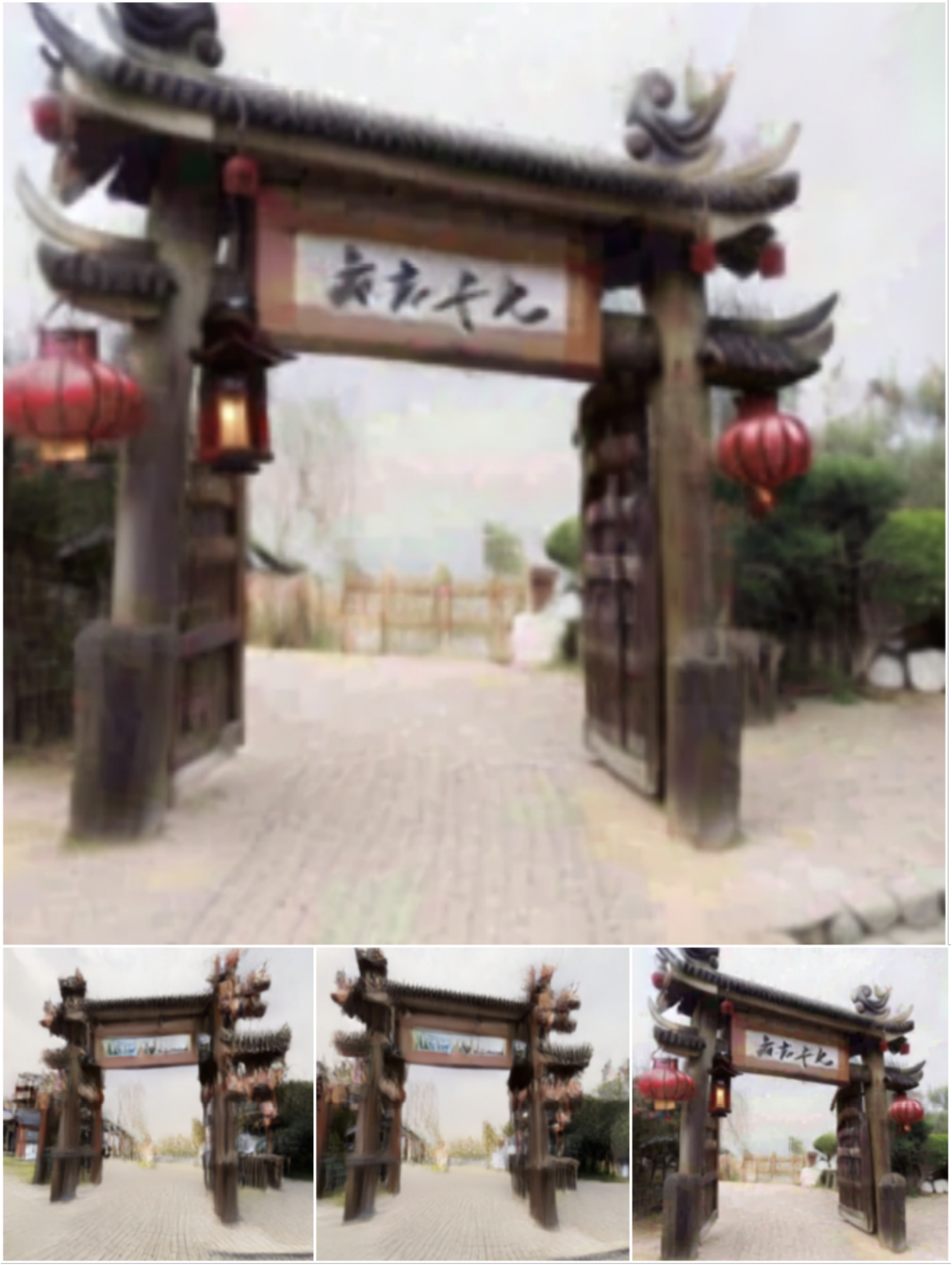}
    & \includegraphics[width=\fourwide]{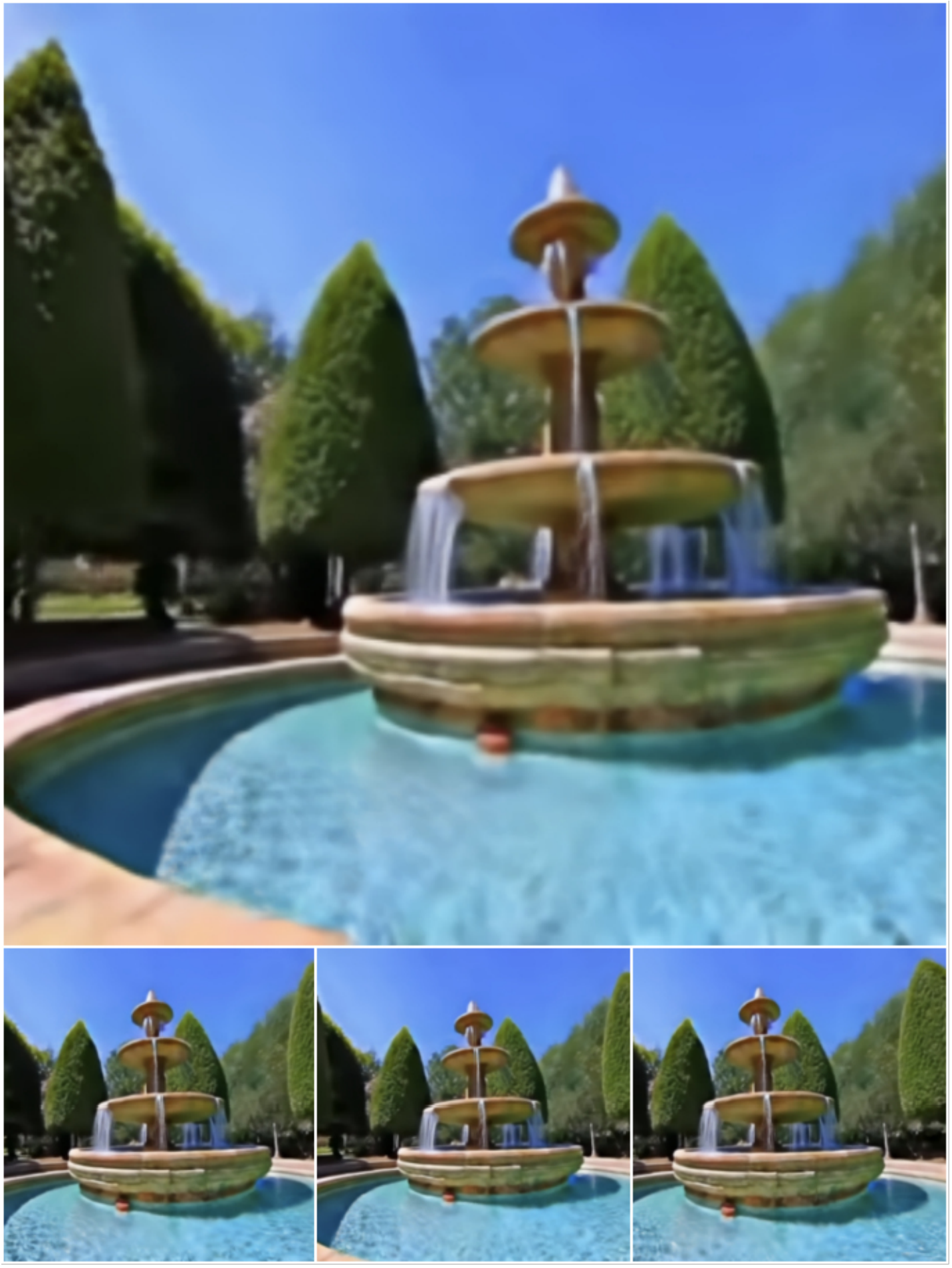}
    \\
    \end{tabular}
    \caption{{\bf Text-to-3D scene generation results of different methods}.}
    \label{fig:qualitative_t23d}
\end{figure}

We compare our method against several state-of-the-art text-to-3D scene generation approaches, including Director3D~\citep{li2024director3d}, Prometheus~\citep{yang2025prometheus}, SplatFlow~\citep{go2025splatflow}, and VideoRFSplat~\citep{go2025splatflow}. 
A qualitative comparison is presented in Fig.~\ref{fig:qualitative_t23d}.
Director3D relies on per-scene refinement, which frequently introduces blurry and wave-like artifacts in the generated results.
In contrast, our model produces accurate objects with fine-grained details, such as animal fur, while preserving realistic backgrounds.
Prometheus does not utilize refinement, and due to the inherent inconsistency of its MV-oriented pipeline, the generated scenes are often blurry and may exhibit incorrect object geometries (\eg, chair legs).
Our approach, however, is capable of generating structurally rich and precise objects in complex scenes, even under large camera movements.
SplatFlow and VideoRFSplat also suffer from blurry artifacts and have difficulty reproducing fine details, such as those found in floors and grass.
In comparison, our model generates realistic details while maintaining semantic consistency with the input text prompt.

\begin{table}[t]
    \centering
    \resizebox{\textwidth}{!}{
    \begin{tabular}{l|ccccc|ccccc|ccccc|c}
    \toprule[0.14em]
    \multirow{3}{*}{\bf Method} & \multicolumn{5}{c|}{\bf T3Bench-200} & \multicolumn{5}{c|}{\bf DL3DV-200} & \multicolumn{5}{c|}{\bf WorldScore-200} & \multirow{3}{*}{\makecell{Time\\Cost}} \\
    & \makecell{Q-Align\\IQA} & \makecell{Q-Align\\IAA} & \makecell{CLIP\\IQA+} & \makecell{CLIP\\Aesthetic} & \makecell{CLIP\\Score} 
    & \makecell{Q-Align\\IQA} & \makecell{Q-Align\\IAA} & \makecell{CLIP\\IQA+} & \makecell{CLIP\\Aesthetic} & \makecell{CLIP\\Score} 
    & \makecell{Q-Align\\IQA} & \makecell{Q-Align\\IAA} & \makecell{CLIP\\IQA+} & \makecell{CLIP\\Aesthetic} & \makecell{CLIP\\Score} 
    &  \\
    \midrule
    Director3D 
    & \cellcolor{orange!50}3.24 & \cellcolor{orange!50}1.95 & \cellcolor{orange!50}0.43 & \cellcolor{orange!50}4.70 & \cellcolor{red!50}\bf 27.84 
    & \cellcolor{orange!50}2.51 & \cellcolor{yellow!25}1.78 & \cellcolor{yellow!25}0.34 & \cellcolor{yellow!25}4.55 & \cellcolor{orange!50}26.12 
    & \cellcolor{orange!50}2.55 & \cellcolor{yellow!25}2.47 & \cellcolor{yellow!25}0.35 & \cellcolor{orange!50}5.32 & \cellcolor{orange!50}29.05 
    & \cellcolor{yellow!25}7 min \\
    
    Promtheus
    & \cellcolor{yellow!25}2.34 & \cellcolor{yellow!25}1.92 & \cellcolor{yellow!25}0.34 & \cellcolor{red!50}\bf 4.76 & \cellcolor{yellow!25}24.85 
    & \cellcolor{yellow!25}2.07 & \cellcolor{orange!50}1.99 & \cellcolor{orange!50}0.35 & \cellcolor{orange!50}4.69 & \cellcolor{yellow!25}23.49 
    & \cellcolor{yellow!25}2.45 & \cellcolor{red!50}\bf 2.94 & \cellcolor{orange!50}0.37 & \cellcolor{red!50}\bf 5.65 & \cellcolor{yellow!25}28.07 
    & \cellcolor{orange!50}15 sec \\
    
    Ours
    & \cellcolor{red!50}\bf 4.12 & \cellcolor{red!50}\bf 2.26 & \cellcolor{red!50}\bf 0.54 & \cellcolor{yellow!25}4.49 & \cellcolor{orange!50}27.68 
    & \cellcolor{red!50}\bf 3.96 & \cellcolor{red!50}\bf 2.27 & \cellcolor{red!50}\bf 0.50 & \cellcolor{red!50}\bf 4.77 & \cellcolor{red!50}\bf 27.63 
    & \cellcolor{red!50}\bf 3.76 & \cellcolor{orange!50}2.55 & \cellcolor{red!50}\bf 0.49 & \cellcolor{yellow!25}5.08 & \cellcolor{red!50}\bf 29.13 
    & \cellcolor{red!50}\bf 9 sec \\
    \bottomrule[0.14em]
    \end{tabular}
    }
    \caption{{\bf Quantitative comparison on text-to-3D scene generation.} Cell background colors indicate the method is the {\colorbox{red!50}{best}}, {\colorbox{orange!50}{second best}}, or {\colorbox{yellow!50}{third best}} on this metric.}
    \label{tab:quantitative_t23d}
\end{table}

We further perform a comprehensive quantitative evaluation for this task.
Specifically, we sample 600 text prompts from T3Bench~\citep{he2023t3bench}, DL3DV~\citep{ling2024dl3dv}, and WorldScore~\citep{duan2025worldscore}, covering object-centric and general scenes.
As all compared methods are based on 3D Gaussian representations, metrics related to camera control and 3D consistency are not applicable in this setting.
Accordingly, we concentrate on the quality evaluation metrics utilized, including CLIP IQA+~\citep{wang2023exploring}, CLIP Aesthetic~\citep{clipaesthetic}, the text-image alignment score (CLIP Score)~\citep{hessel2021clipscore}, as well as the latest LMM-based Q-Align~\citep{wu2024qalign} image quality metric.
The quantitative results are summarized in Tab.~\ref{tab:quantitative_t23d}.
It is evident that our model achieves superior performance on the majority of quality evaluation metrics.
For CLIP-Aesthetic, we note that this metric sometimes favor smooth outputs, which may not always align with the detailed and realistic results produced by our method.
Our method also attains the highest CLIP Score for two subsets, demonstrating the strong text alignment ability of our method.
In addition, we report the average time required to generate a single scene for each method on a single H20 GPU.
Our method demonstrates a substantial speed advantage over other approaches.
Remarkably, this efficiency is maintained even when our method produces results with higher resolution and a greater number of frames.
In addition, our approach leverages a unified model that seamlessly handles both image-to-3D and text-to-3D tasks without requiring separate training processes.
This unified framework not only simplifies the overall workflow but also substantially reduces the training cost.

\subsection{Comparison on WorldScore Benchmark}

\begin{figure}[t]
    \newcommand{\leftwide}{7.78cm}
    \newcommand{\rightwide}{5.33cm}
    
    \begin{tabular}{c@{\,\,}|c@{\,\,}}
    Ours
    & WonderWorld
    \\
    \includegraphics[width=\leftwide]{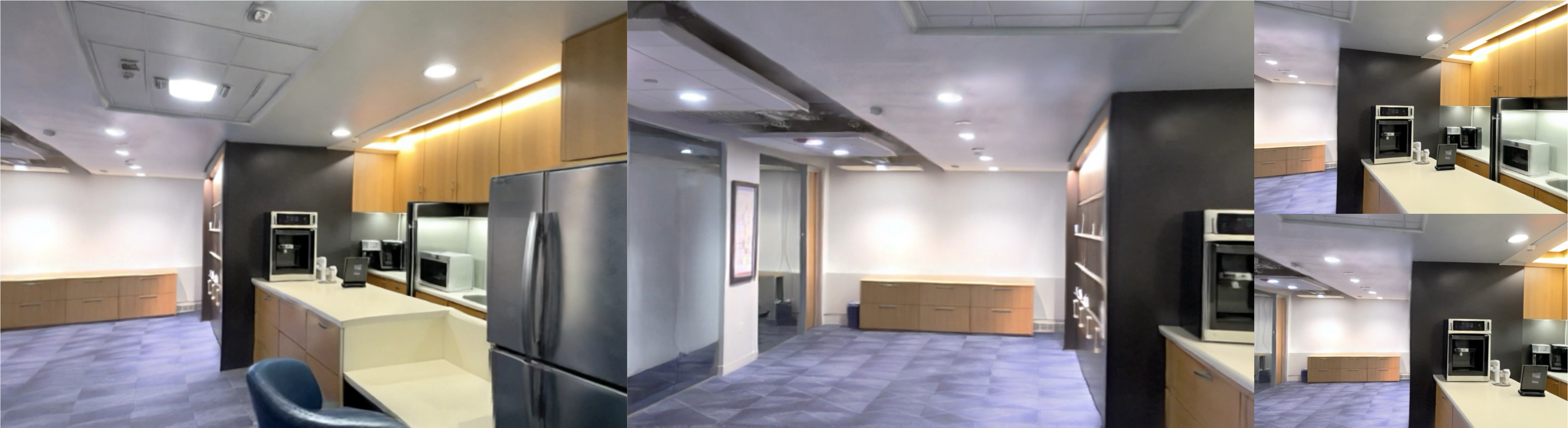}
    & \includegraphics[width=\rightwide]{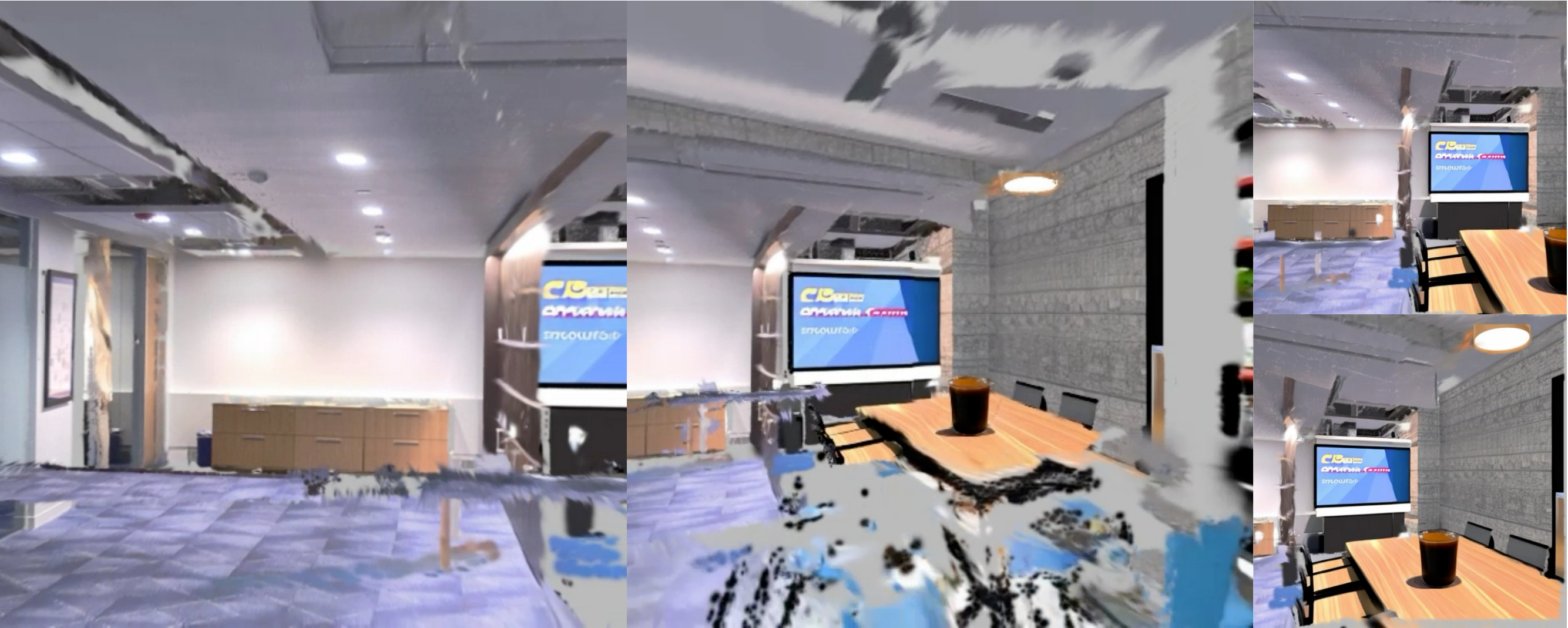}
    \\
    
    & LucidDreamer
    \\
    \includegraphics[width=\leftwide]{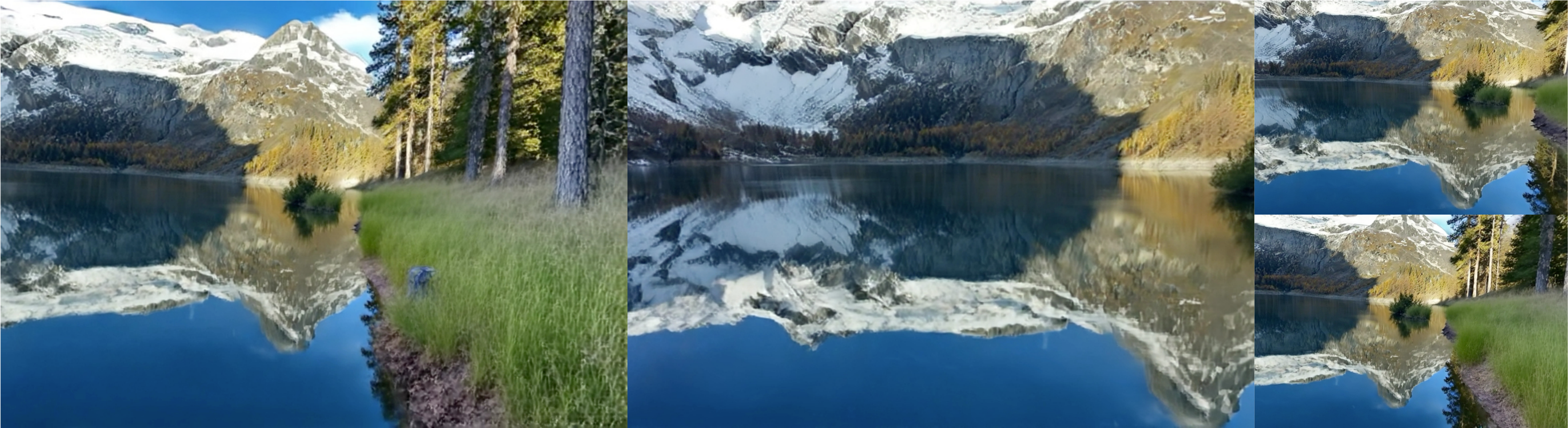}
    & \includegraphics[width=\rightwide]{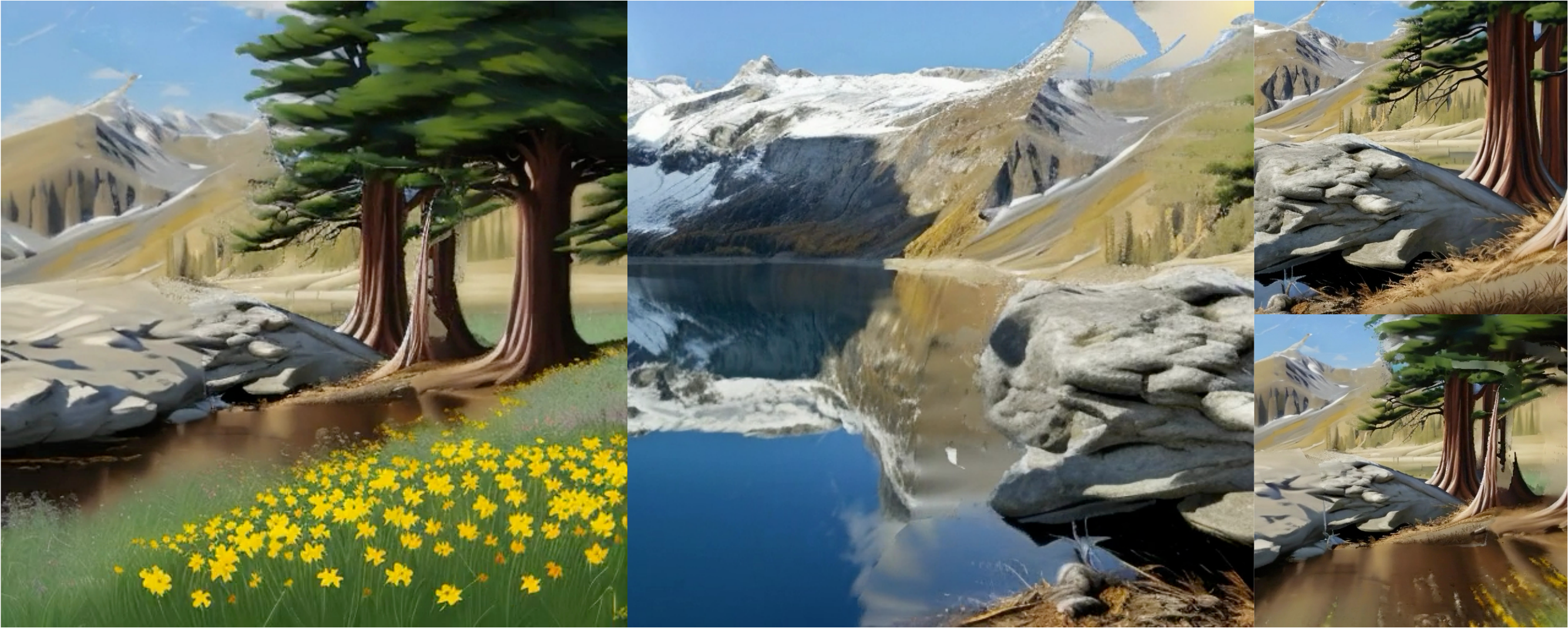}
    \\
    
    & WonderJourney
    \\
    \includegraphics[width=\leftwide]{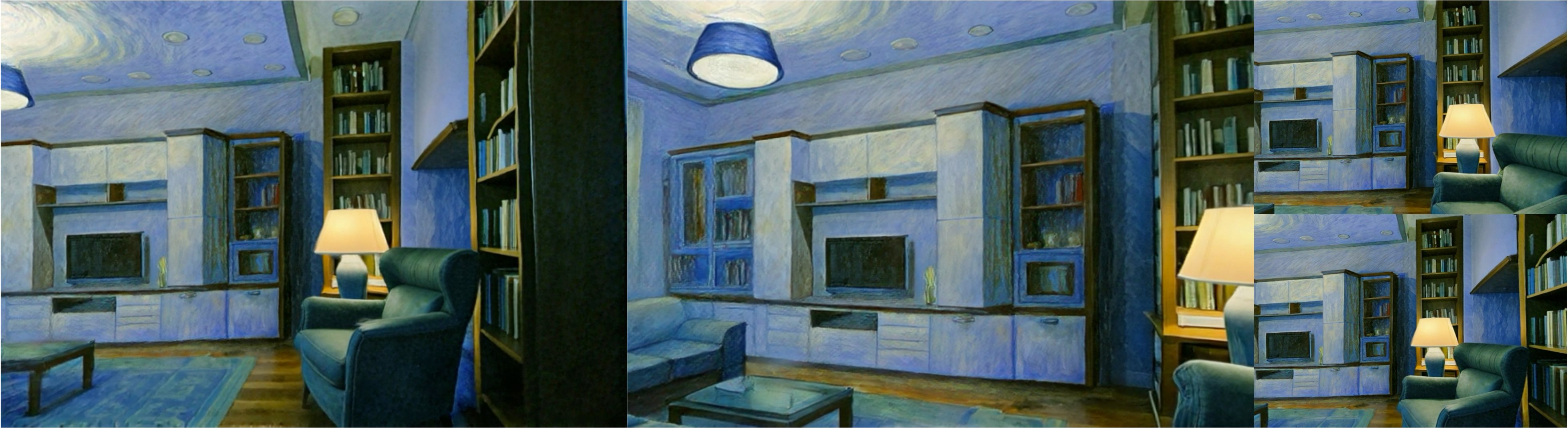}
    & \includegraphics[width=\rightwide]{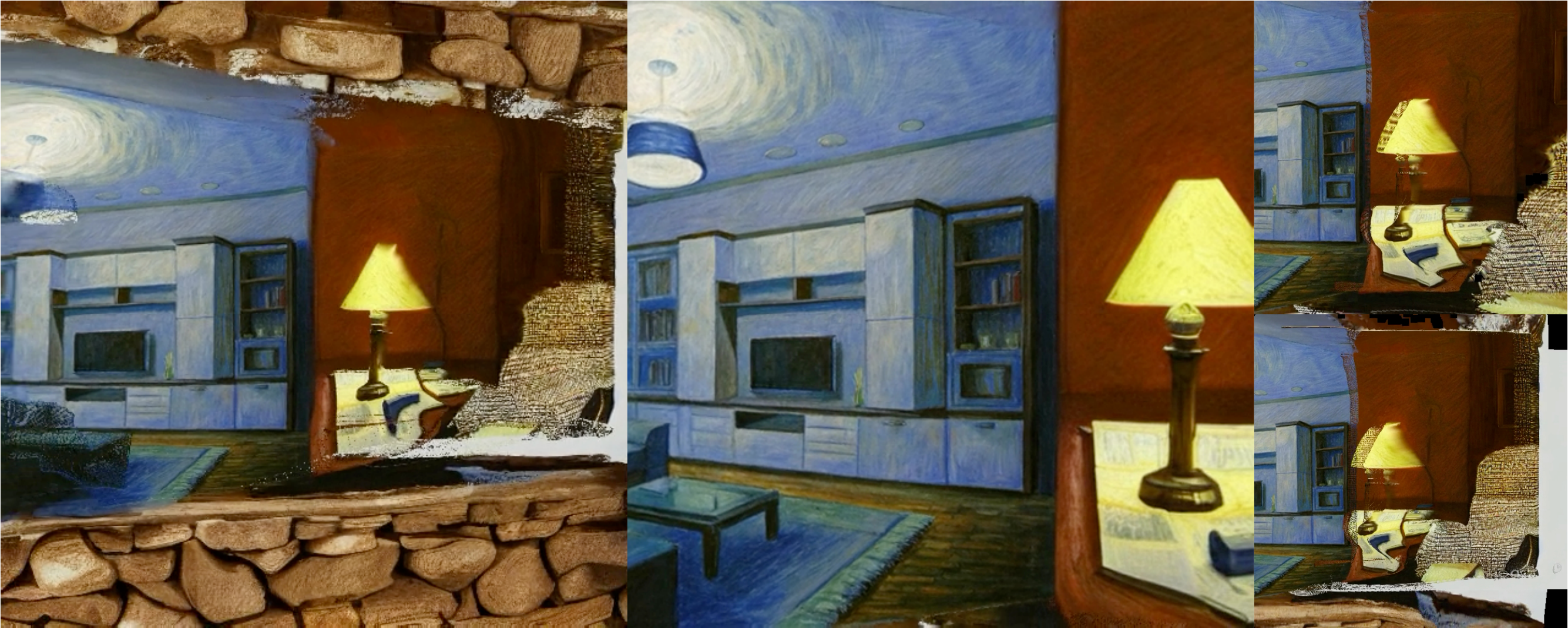}
    \\
    \end{tabular}
    \caption{{\bf 3D scene generation results of different methods on WorldScore benchmark.}}
    \label{fig:qualitative_worldscore}
\end{figure}

\begin{table}[t]
    \centering
    \resizebox{\textwidth}{!}{
    \begin{tabular}{l|cccccc|c|c}
    \toprule[0.14em]
    \multirow{2}{*}{\bf Method} & \multirow{2}{*}{\makecell{3D\\Consistency}} & \multirow{2}{*}{\makecell{Photometric\\Consistency}} & \multirow{2}{*}{\makecell{Object\\Control}} & \multirow{2}{*}{\makecell{Content\\Alignment}} & \multirow{2}{*}{\makecell{Style\\Consistency}} & \multirow{2}{*}{\makecell{Subjective\\Quality}} & \multirow{2}{*}{\makecell{\bf Average}} & \multirow{2}{*}{\makecell{Time\\Cost}} \\
    & & & & & & & & \\  %
    \midrule
    WonderJourney  & 80.60 & 79.03 & 34.81 & 38.37 & \cellcolor{yellow!50}67.52 & \cellcolor{red!50}\bf 61.49 & 60.30 & \cellcolor{yellow!50}6 min \\
    LucidDreamer   & \cellcolor{red!50}\bf 90.37 & \cellcolor{red!50}\bf 90.20 & \cellcolor{yellow!50}43.48 & \cellcolor{red!50}\bf 59.41 & 66.41 & \cellcolor{yellow!50}48.02 & \cellcolor{yellow!50}66.32 & \cellcolor{yellow!50}6 min \\
    WonderWorld    & \cellcolor{orange!50}86.91 & \cellcolor{yellow!50}85.56 & \cellcolor{red!50}\bf 52.09 & \cellcolor{orange!50}56.82 & \cellcolor{orange!50}75.92 & 41.28 & \cellcolor{orange!50}66.43 & \cellcolor{orange!50}10 sec \\
    Ours           & \cellcolor{yellow!50}85.87 & \cellcolor{orange!50}86.72 & \cellcolor{orange!50}49.61 & \cellcolor{yellow!50}53.96 & \cellcolor{red!50}\bf 81.52 & \cellcolor{orange!50}54.63 & \cellcolor{red!50}\bf 68.72 & \cellcolor{red!50}\bf 9 sec \\
    \bottomrule[0.14em]
    \end{tabular}
    }
    \caption{{\bf Quantitative comparison on WorldScore benchmark.} Note that the time cost of the baselines is tested on 1$\times$ H100 GPU, while our time cost is tested on 1$\times$ H20 GPU.}
    \label{tab:quantitative_worldscore}
\end{table}

We further conduct a comprehensive evaluation on the recent WorldScore~\citep{duan2025worldscore} benchmark.
The static subset of WorldScore comprises 2,000 test examples, encompassing a diverse array of worlds with varying styles, scenarios, and objects.
Each test case provides an input image, a text prompt, and a camera trajectory as conditions for generation.
The evaluation protocol is designed to assess two primary aspects of world generation: controllability and quality.
For baselines, we select three state-of-the-art 3D generation methods:
WonderJourney~\citep{yu2024wonderjourney}, which iteratively completes novel-view images and depth maps based on point clouds;
LucidDreamer~\citep{chung2023luciddreamer}, which also performs iterative novel view completion but utilizes 3DGS for rendering;
and WonderWorld~\citep{yu2025wonderworld}, which improves generation quality through the use of layered Gaussian surfels.
Since our comparison focuses exclusively on 3D generation methods, the ``Camera Control'' metric primarily reflects the robustness of the evaluation protocol for each method, and is thus less informative in this context.
Accordingly, we omit this metric from our comparison.
Additionally, the original WorldScore benchmark evaluates most metrics only on anchor frames, which is suboptimal for 3D world generation tasks that require novel view synthesis.
To ensure a fairer comparison, we re-evaluate these metrics by randomly sampled frames within specific intervals.
Qualitative and quantitative comparisons are shown in Fig.~\ref{fig:qualitative_worldscore} and Tab.~\ref{tab:quantitative_worldscore}, respectively. 
Our method achieves the highest average score and the fastest inference speed among all compared approaches. 
In particular, our model achieves the best ``Style Consistency'' and secures the second place in ``Photometric Consistency'', ``Object Control'', and ``Subjective Quality'', reflecting a well-balanced and robust capability across controllability and quality. 
While our method yields relatively lower scores in ``3D Consistency'' and ``Content Alignment'', these results can be attributed to methodological differences: for ``3D Consistency'', all baselines utilize monocular depth estimation models that are closely aligned with the evaluation protocol, whereas our approach relies solely on RGB supervision without explicit depth guidance; for ``Content Alignment'', our method does not directly manipulate the anchor frame content, in contrast to the baselines. 
Qualitative analysis further reveals that baseline methods frequently exhibit unnatural transitions, discontinuous content, and visible holes in the generated scenes, which may not be fully reflected by the current metrics. 
Overall, our approach demonstrates superior consistency and faithful generation over existing methods.

\subsection{Ablation Study}

\begin{figure}[t]
    \newcommand{\sevenwide}{2.15cm}
    
    \begin{tabular}{c@{\,\,}c@{\,\,}c@{\,\,}c@{\,\,}c@{\,\,}c@{\,\,}}
    \textit{w/} MV-Diff
    & \textit{w/} 3D-Diff
    & \textit{w/} MV-Dist
    & \textit{w/o} CMC
    & \textit{w/o} OOD
    & Full model
    \\
    \includegraphics[width=\sevenwide]{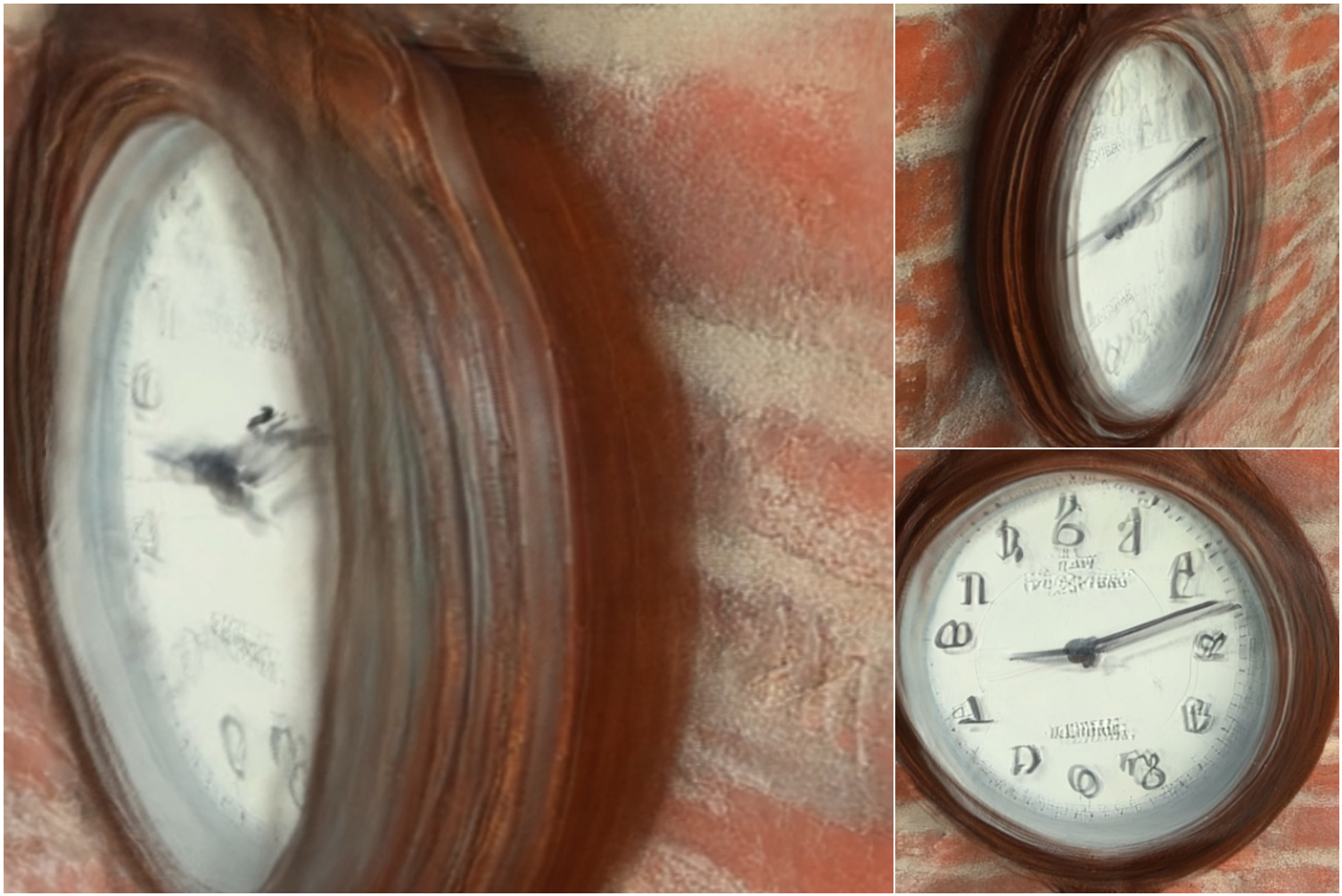}
    & \includegraphics[width=\sevenwide]{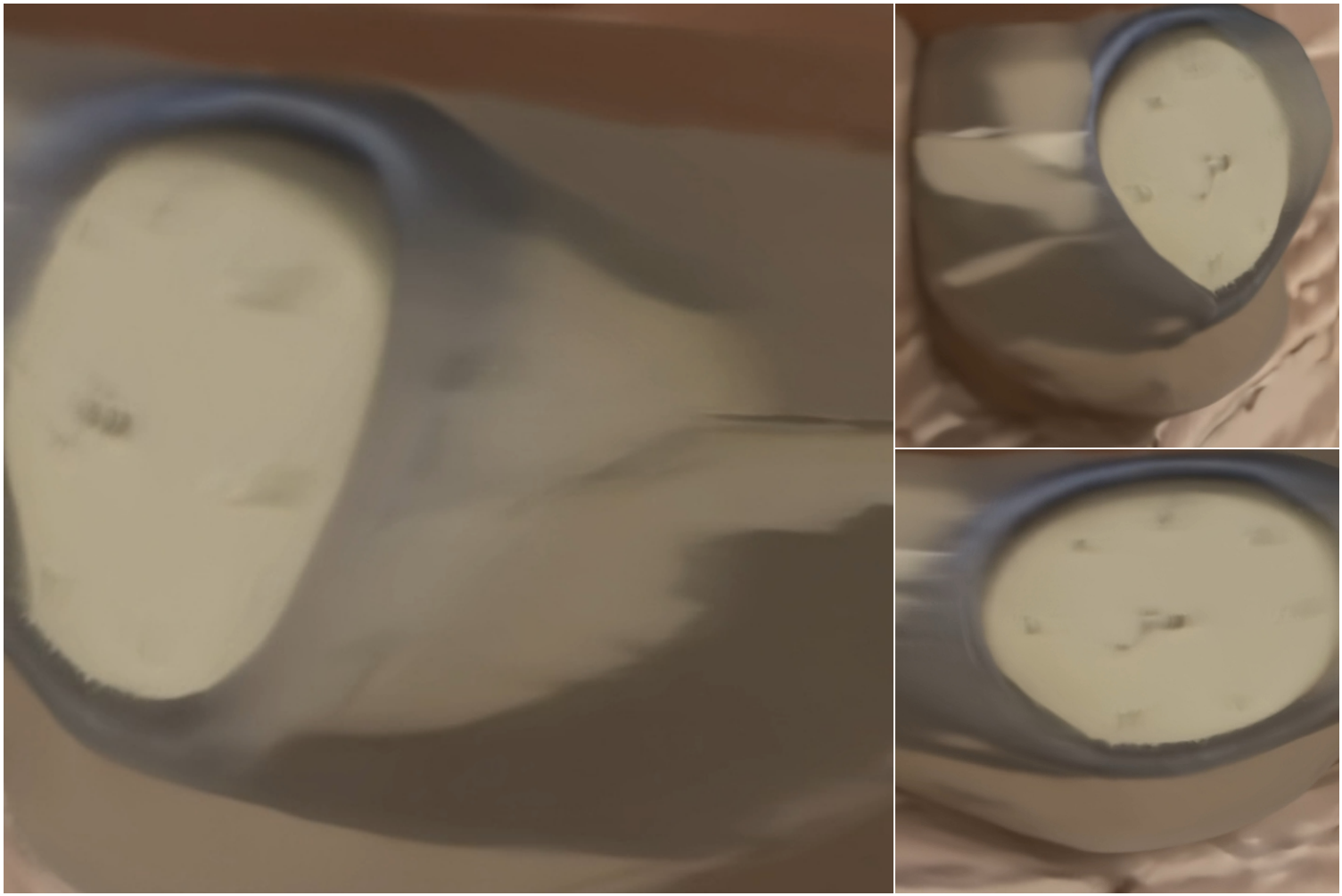}
    & \includegraphics[width=\sevenwide]{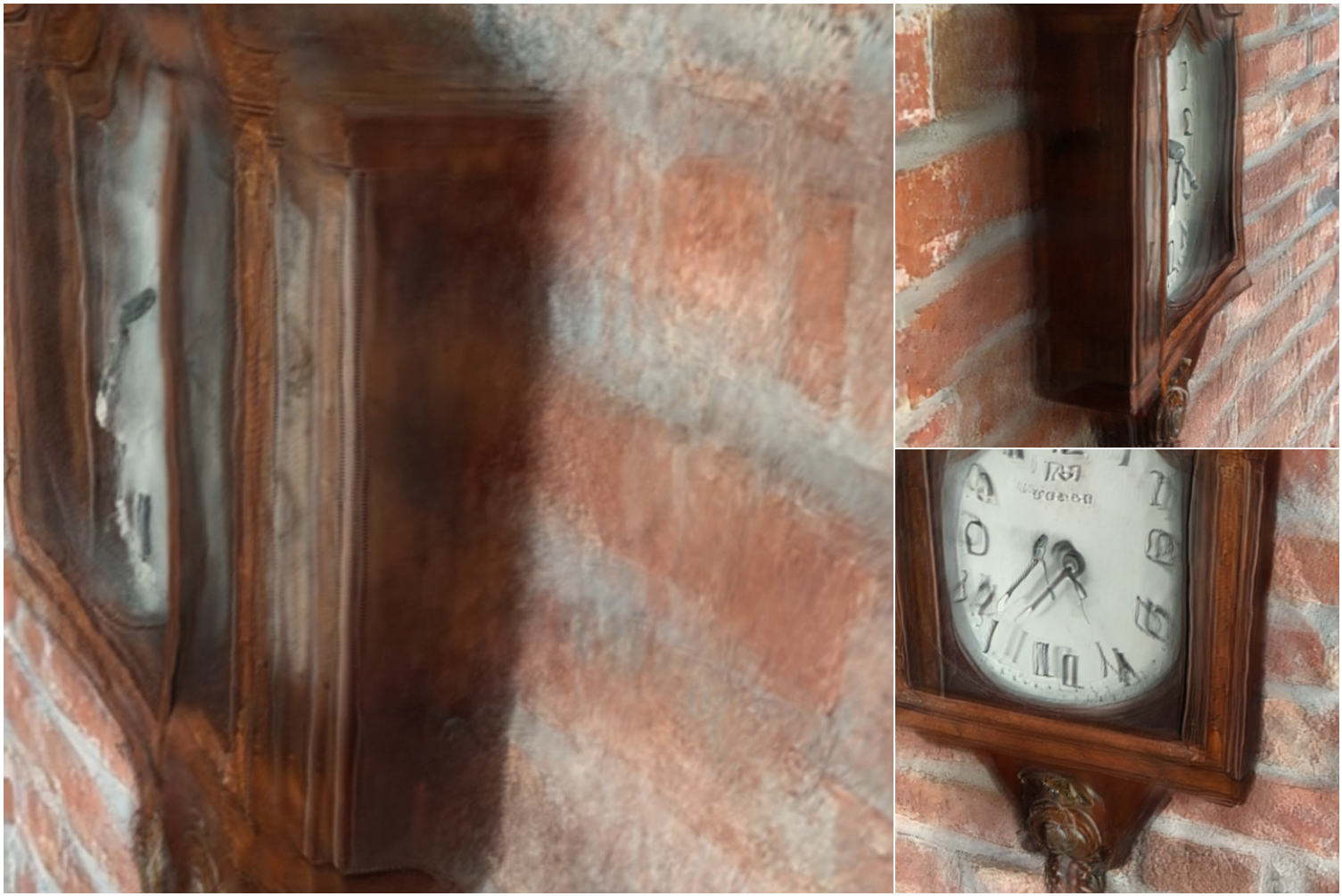}
    & \includegraphics[width=\sevenwide]{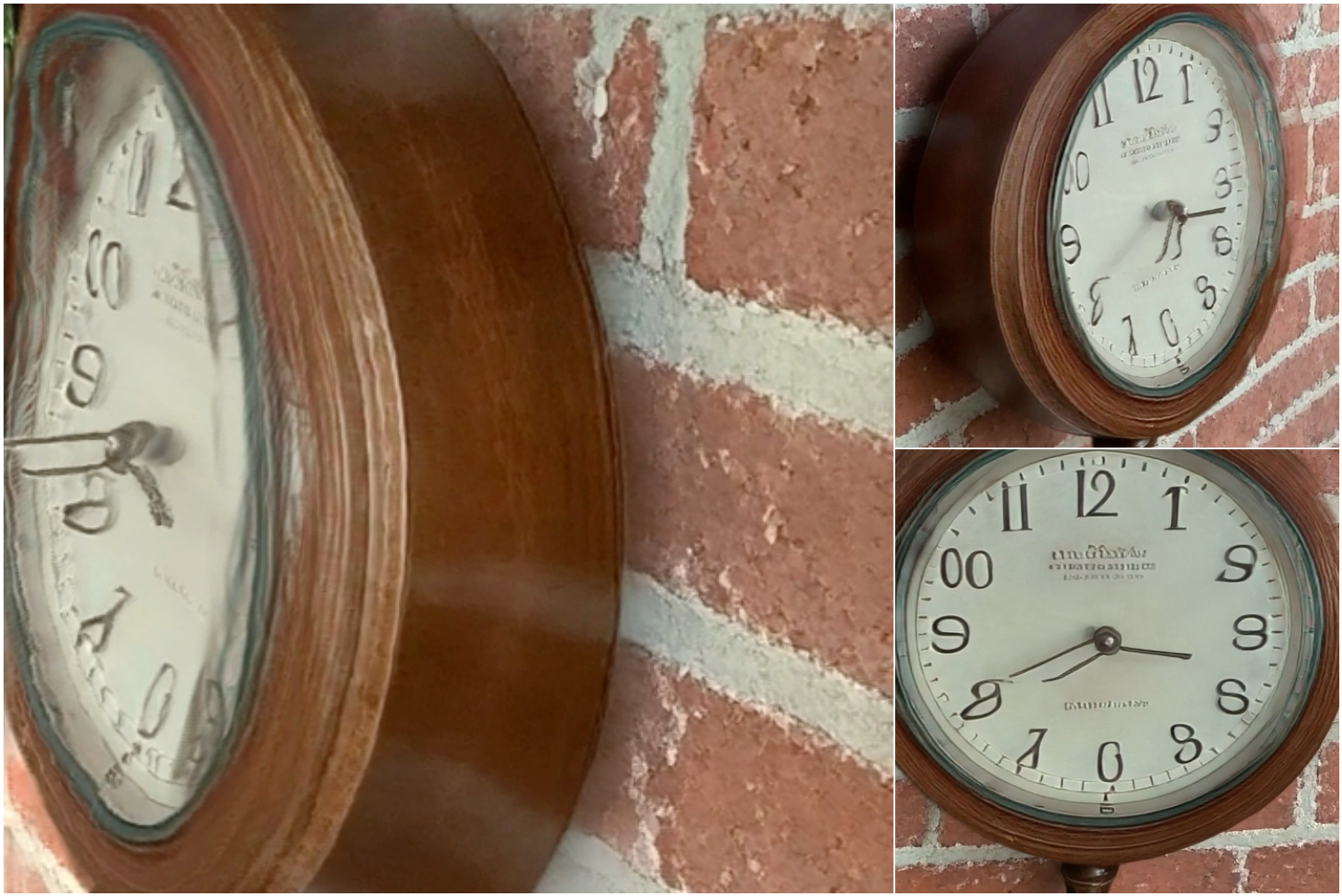} 
    & \includegraphics[width=\sevenwide]{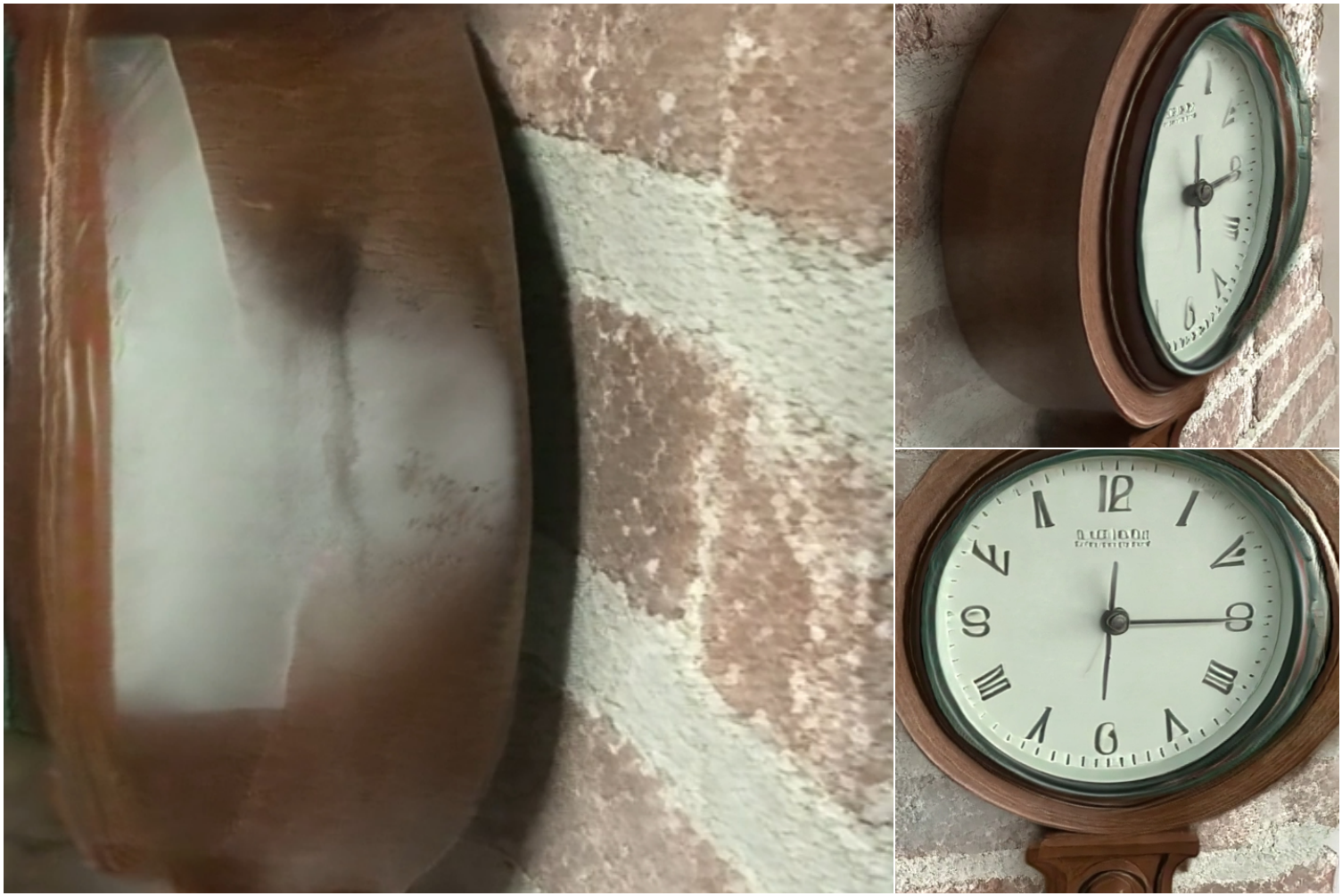}
    & \includegraphics[width=\sevenwide]{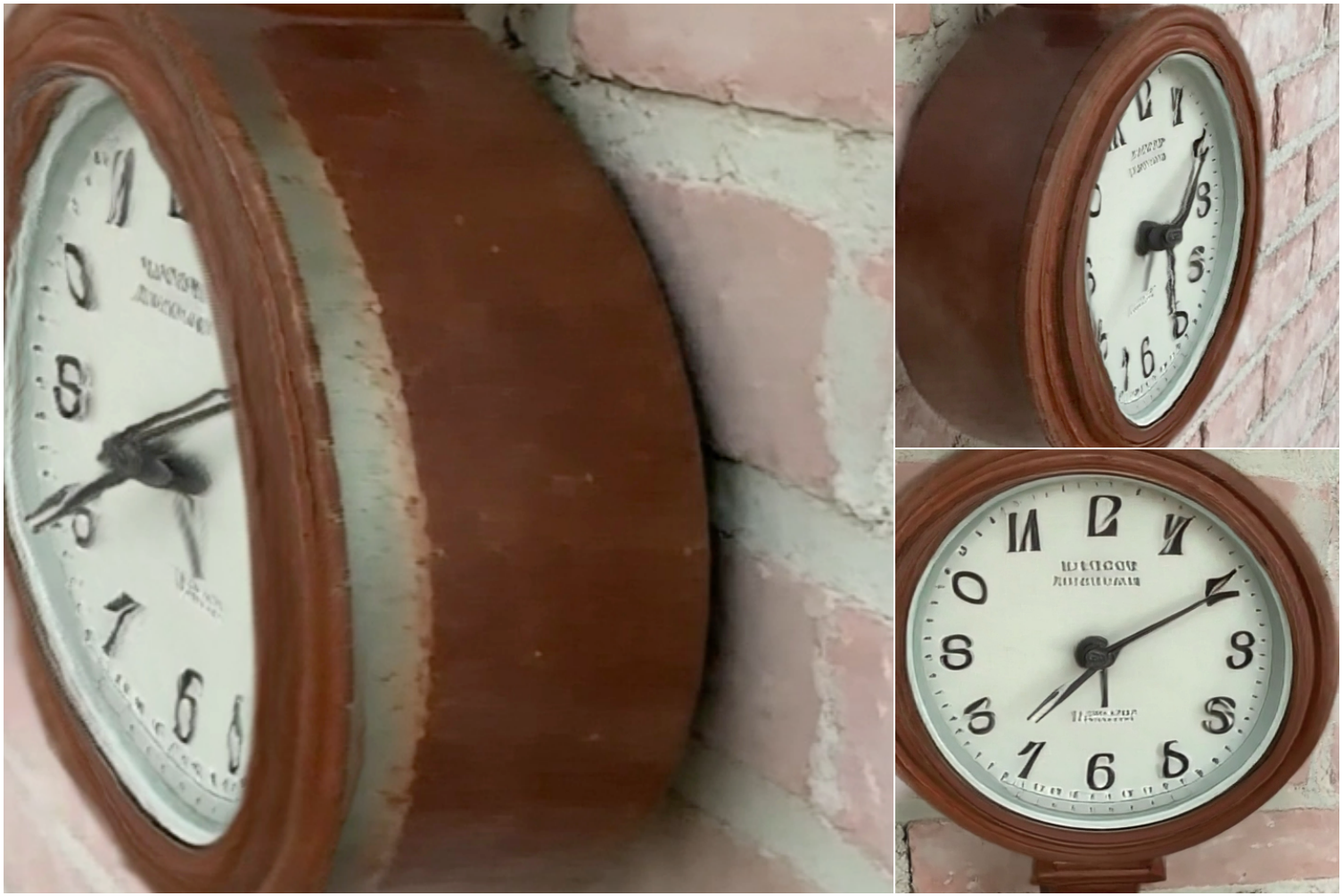} 
    \\
    \includegraphics[width=\sevenwide]{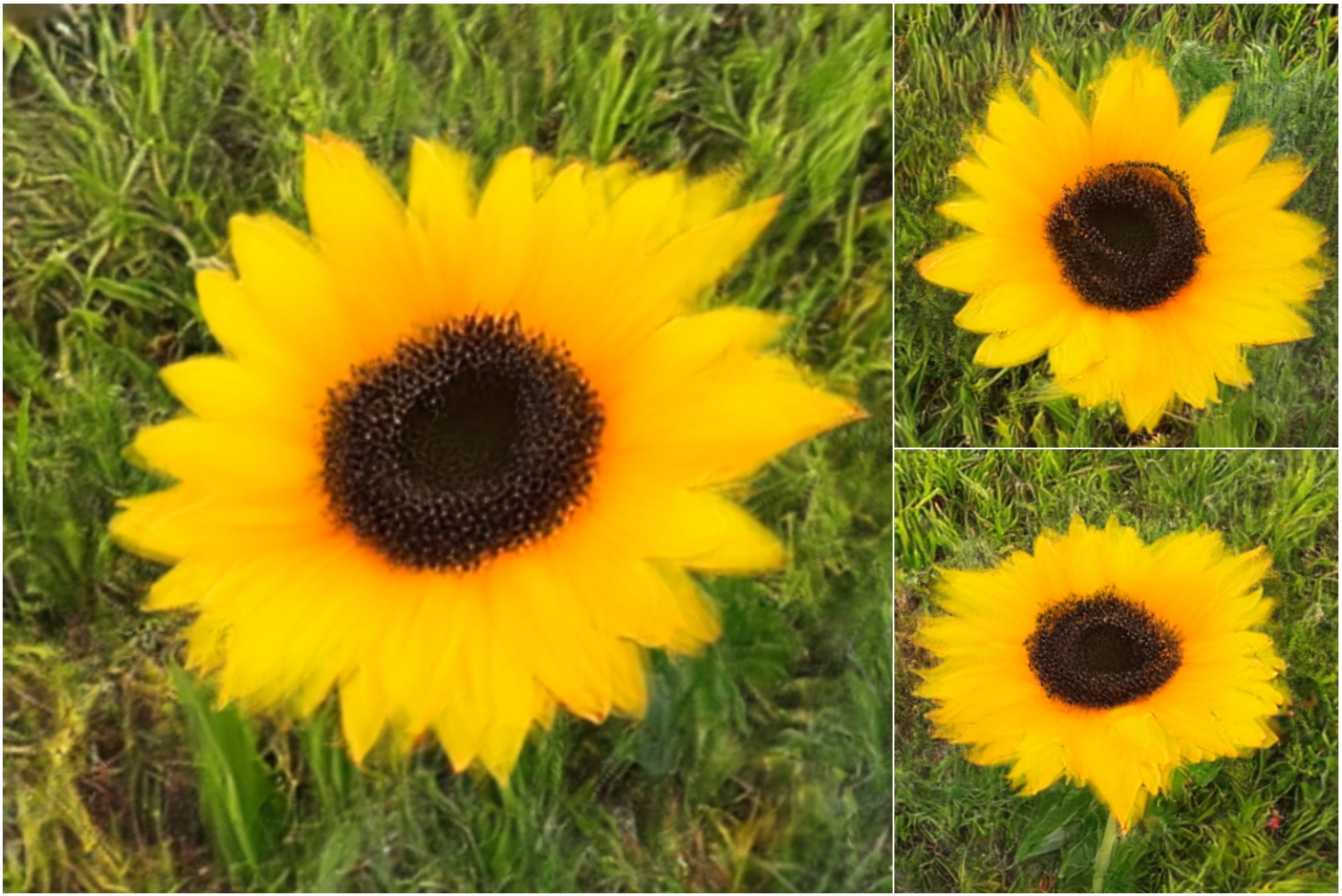}
    & \includegraphics[width=\sevenwide]{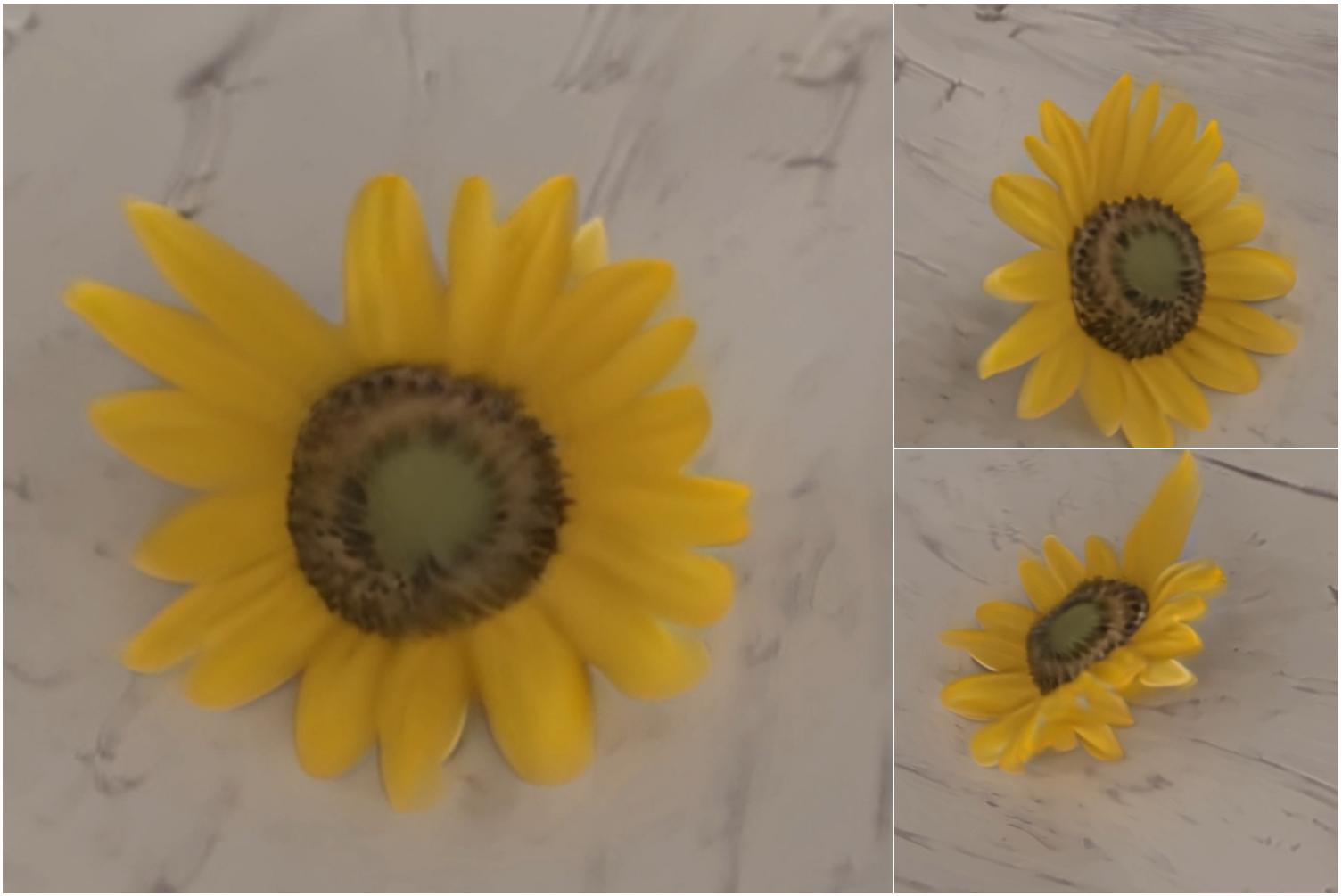}
    & \includegraphics[width=\sevenwide]{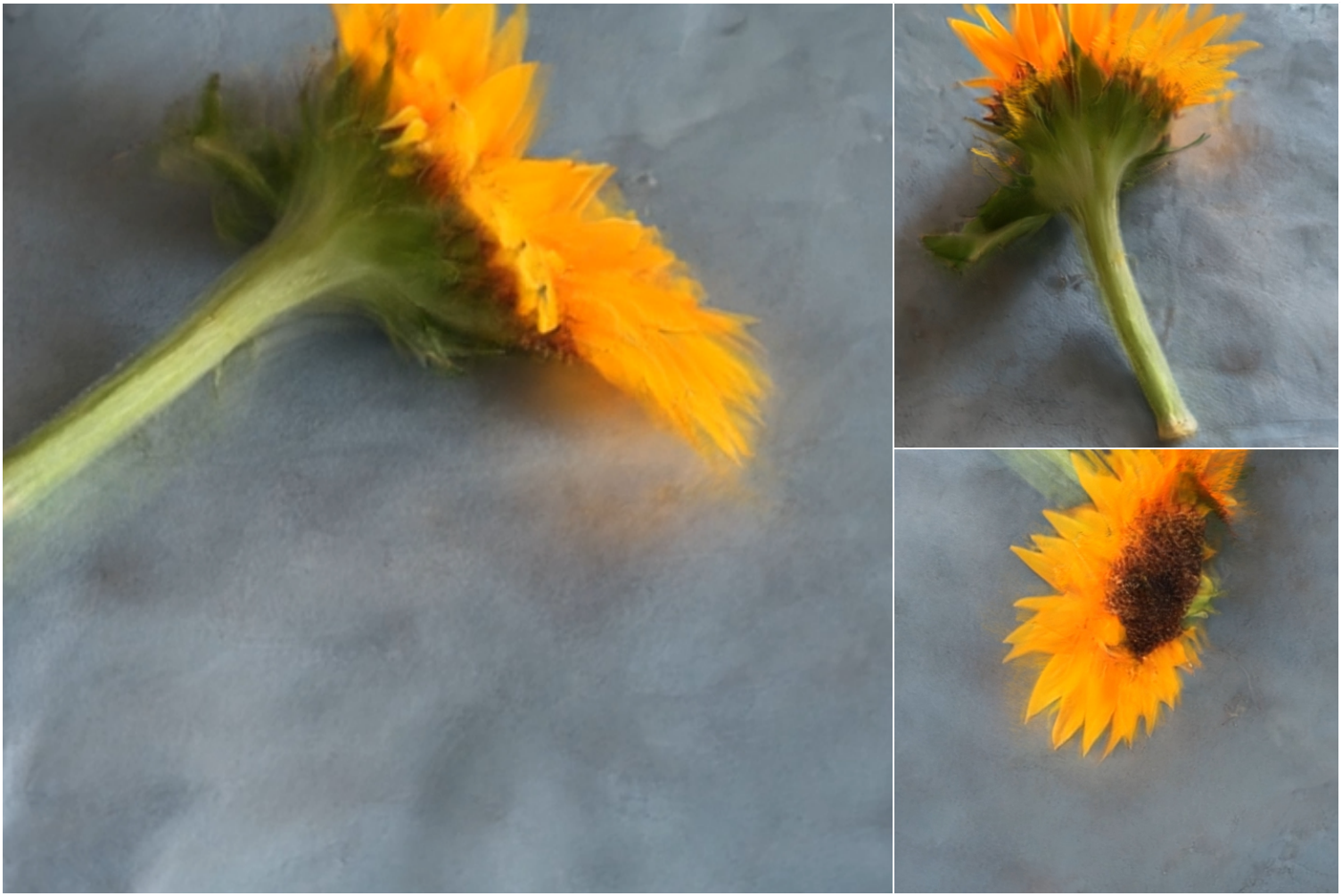}
    & \includegraphics[width=\sevenwide]{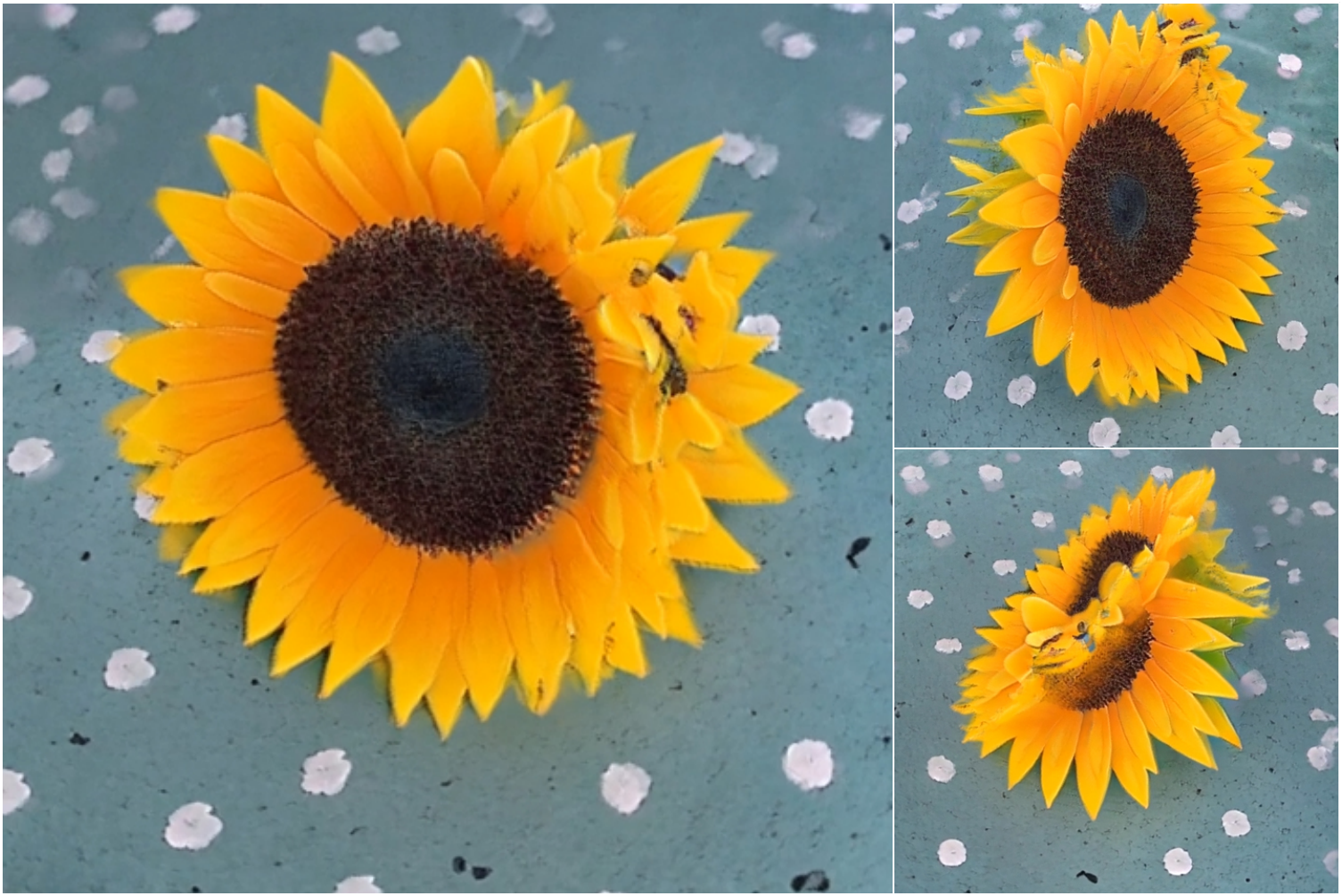} 
    & \includegraphics[width=\sevenwide]{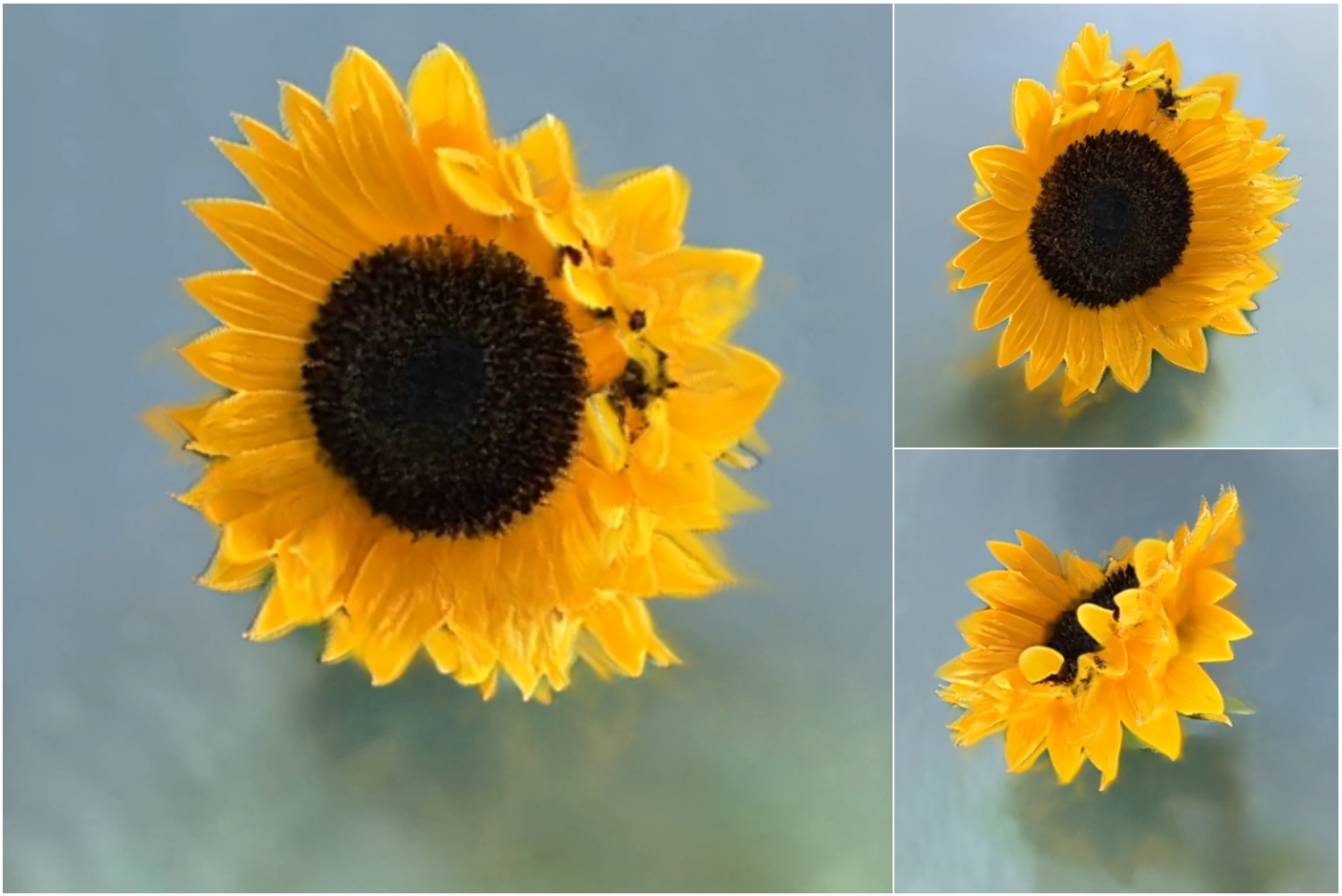}
    & \includegraphics[width=\sevenwide]{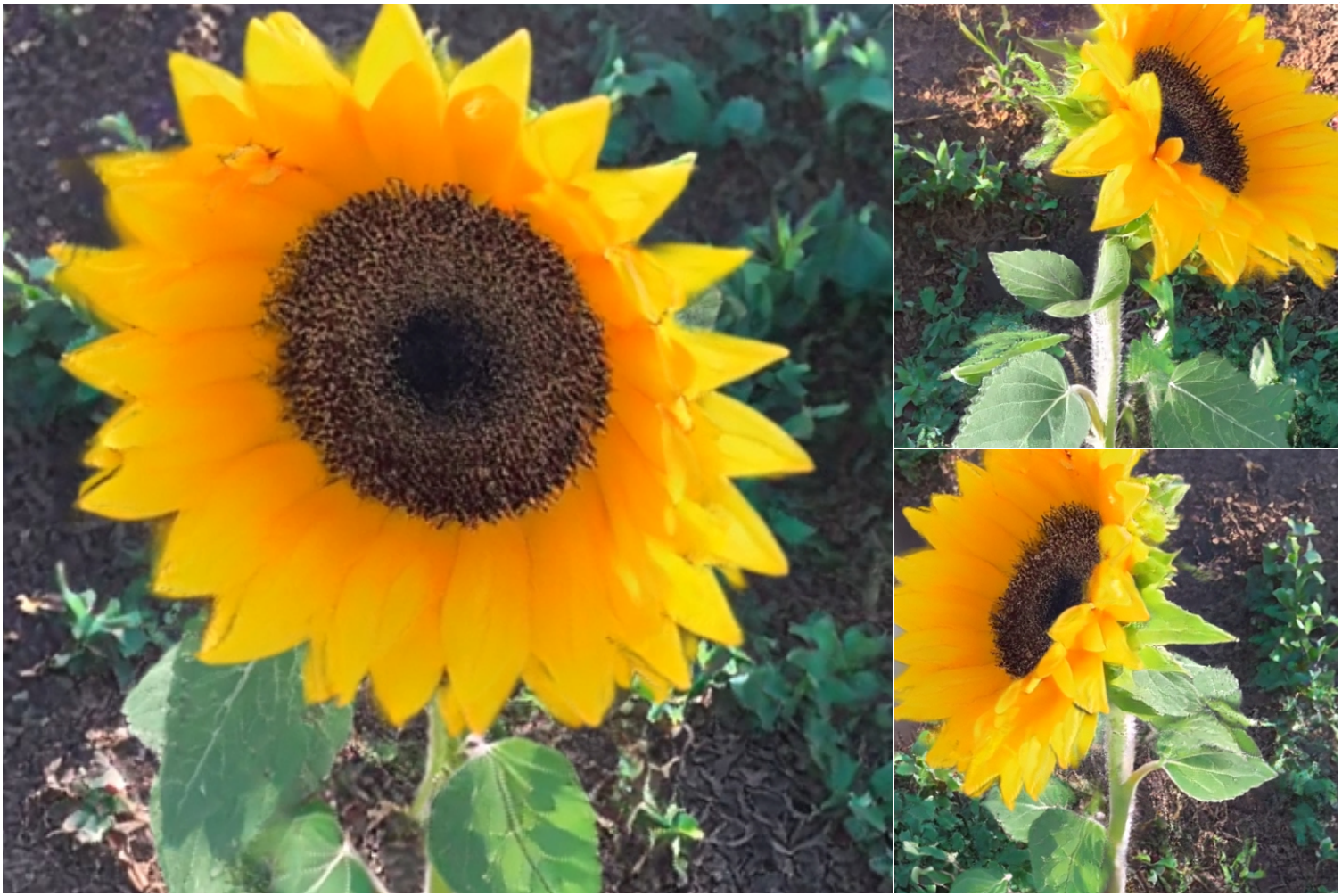} 
    \\
    \end{tabular}
    \caption{{\bf Qualitative ablation studies.} Prompts: (Top) {\fontfamily{ppl}\selectfont ``A vintage clock hanging on a brick wall''}; (Bottom) {\fontfamily{ppl}\selectfont ``A bright sunflower in a field''}. }
    \label{fig:qualitative_ablation}
\end{figure}

In Fig.~\ref{fig:motivation}, we show the generation results of various ablation models for image-to-3D scene generation. 
The outcomes align well with our expectations: both the MV-oriented diffusion model (\textit{w/} MV-Diff) and the MV-oriented distillation model (\textit{w/} MV-Dist) exhibit noisy 3D reconstruction due to multi-view inconsistency, while the 3D-oriented diffusion model (\textit{w/} 3D-Diff) produces blurry visual results.
To further validate the effectiveness of each proposed strategy, we conduct more comprehensive ablation studies on text-to-3D scene generation. 
Quantitative and qualitative results are summarized in Tab.~\ref{tab:quantitative_ablation} and Fig.~\ref{fig:qualitative_ablation}, respectively.
Consistently, the first three ablation models continue to demonstrate worse visual quality and weaker text alignment.
The model without cross-mode consistency loss (\textit{w/o} CMC) achieves competitive, and in some cases superior, scores on most quantitative metrics compared to our full model. However, qualitative analysis reveals that this model is susceptible to floating and duplicated artifacts.
The model without out-of-distribution data (\textit{w/o} OOD) is more prone to semantic misalignment (\eg, ``field'') and exhibits a drop in quantitative text alignment metrics. This issue is exacerbated on T3Bench and WorldScore, which differ in distribution from the original multi-view data, highlighting the importance of incorporating OOD data to improve generalization.

\begin{table}[t]
    \centering
    \resizebox{\textwidth}{!}{ %
    \begin{tabular}{l|ccccc|ccccc|ccccc}
    \toprule[0.14em]
    \multirow{3}{*}{\bf} & \multicolumn{5}{c|}{\bf T3Bench-200} & \multicolumn{5}{c|}{\bf DL3DV-200} & \multicolumn{5}{c}{\bf WorldScore-200} \\
    & \makecell{Q-Align\\IQA} & \makecell{Q-Align\\IAA} & \makecell{CLIP\\IQA+} & \makecell{CLIP\\Aesthetic} & \makecell{CLIP\\Score} 
    & \makecell{Q-Align\\IQA} & \makecell{Q-Align\\IAA} & \makecell{CLIP\\IQA+} & \makecell{CLIP\\Aesthetic} & \makecell{CLIP\\Score} 
    & \makecell{Q-Align\\IQA} & \makecell{Q-Align\\IAA} & \makecell{CLIP\\IQA+} & \makecell{CLIP\\Aesthetic} & \makecell{CLIP\\Score} \\
    \midrule
    \textbf{A} & 3.11 & 2.03 & 0.41 & 4.36 & 25.34 & 2.64 & \cellcolor{yellow!25}2.09 & 0.39 & 4.60 & 24.49 & 2.48 & 2.10 & 0.35 & 4.78 & 27.40 \\
    \textbf{B} & 2.61 & 1.68 & 0.37 & 4.11 & 22.92 & 2.71 & 1.96 & 0.40 & 4.54 & 22.71 & 2.74 & 2.16 & 0.33 & 4.83 & 26.11 \\
    \textbf{C} & \cellcolor{yellow!25}3.46 & 2.12 & 0.45 & 4.42 & 26.95 & 2.99 & 2.05 & \cellcolor{yellow!25}0.42 & 4.57 & 26.41 & 3.06 & 2.18 & 0.42 & 4.92 & 28.71  \\
    \textbf{D} & \cellcolor{red!50}\bf 4.12 & \cellcolor{orange!50}2.31 & \cellcolor{yellow!25}0.52 & \cellcolor{orange!50}4.52 & \cellcolor{orange!50}27.59 & \cellcolor{red!50}\bf 4.02 & \cellcolor{red!50}\bf 2.35 & \cellcolor{red!50}\bf 0.51 & \cellcolor{orange!50}4.80 & \cellcolor{red!50}\bf 27.90 & \cellcolor{red!50}\bf 3.90 & \cellcolor{red!50}\bf 2.71 & \cellcolor{red!50}\bf 0.51 & \cellcolor{red!50}\bf 5.12 & \cellcolor{orange!50}29.09 \\
    \textbf{E} & \cellcolor{orange!50}3.98 & \cellcolor{red!50}\bf 2.50 & \cellcolor{orange!50}0.53 & \cellcolor{red!50}\bf 4.58 & \cellcolor{yellow!25}27.04 & \cellcolor{yellow!25}3.89 & \cellcolor{red!50}\bf 2.35 & \cellcolor{orange!50}0.50 & \cellcolor{red!50}\bf 4.82 & \cellcolor{yellow!25}27.45 & \cellcolor{yellow!25}3.66 & \cellcolor{orange!50}2.56 & \cellcolor{yellow!25}0.47 & \cellcolor{yellow!25}4.95 & \cellcolor{yellow!25}28.76 \\
    \textbf{F} & \cellcolor{red!50}\bf 4.12 & \cellcolor{yellow!25}2.26 & \cellcolor{red!50}\bf 0.54 & \cellcolor{yellow!25}4.49 & \cellcolor{red!50}\bf 27.68 & \cellcolor{orange!50}3.96 & \cellcolor{orange!50}2.27 & \cellcolor{orange!50}0.50 & \cellcolor{yellow!25}4.77 & \cellcolor{orange!50}27.63 & \cellcolor{orange!50}3.76 & \cellcolor{yellow!25}2.55 & \cellcolor{orange!50}0.49 & \cellcolor{orange!50}5.08 & \cellcolor{red!50}\bf 29.13 \\
    \bottomrule[0.14em]
    \end{tabular}
    }
    \caption{{\bf Quantitative ablation studies.} The letters A--F correspond to different model variants: \textbf{(A)} \textit{w/} MV-Diff, \textbf{(B)} \textit{w/} 3D-Diff, \textbf{(C)} \textit{w/} MV-Dist, \textbf{(D)} \textit{w/o} CMC, \textbf{(E)} \textit{w/o} OOD, and \textbf{(F)} Full model.
    }
    \label{tab:quantitative_ablation}
\end{table}

\section{Conclusion}
We propose an efficient yet powerful model for 3D scene generation, named \modelname. At the core of our approach is a novel distillation strategy, which transfers high visual fidelity from a multi-view-oriented diffusion model to a 3D-oriented multi-view generative model endowed with perfect 3D consistency.
To achieve this, we design a dual-mode pre-training phase and a cross-mode post-training phase, and introduce an out-of-distribution data co-training strategy to boost the model's generalization.
Our method achieves state-of-the-art performance on multiple tasks, while offering significant advantages in inference speed.
The efficiency and effectiveness of our approach are well-positioned to advance applications of 3D scene generation.
Future work includes incorporating autoregressive generation~\citep{zhang2024tar3d, zhang2025ar, huang2025self,chen2024diffusion} and extending our framework to dynamic 4D scene generation tasks~\citep{wu20234dgaussians,zhang20244diffusion}.

\bibliography{iclr2025_conference}
\bibliographystyle{iclr2025_conference}

\appendix
\section{Training Details}
\label{app:training_details}

\noindent
\textbf{Architecture configuration. } Our dual-mode multi-view latent diffusion model is initialized with WAN2.2-5B-IT2V~\citep{wan2025wan}.
For both pre-training and post-training, we adopt 24 key frames as input views.
The spatial downsampling factor from image space to latent space is set to 16.
The auxiliary multi-view feature has a channel dimension of 1024.
The discriminator head is a simple CNN with several residual convolutional blocks~\citep{he2016deep}.

\noindent
\textbf{Pre-training configuration. } We set the learning rates for both the transformer and 3DGS decoder to $2 \times 10^{-6}$. We use a weight decay of $1 \times 10^{-6}$ and Adam~\citep{kingma2014adam} optimizer parameters $\beta_1 = 0.9$ and $\beta_2 = 0.95$. The training schedule includes a warm-up phase of 1{,}000 steps, followed by a learning rate decay over 10{,}000 steps, with a total of 20{,}000 training steps. The training takes around 3 days.

\noindent
\textbf{Post-training configuration.} During post-training, the timestep schedule for the few-step generator is set to $\{1000,900,750,500\}$. For each generator update, the fake score network is updated 4 times. The learning rates are set to $1 \times 10^{-6}$ for the generator and $5 \times 10^{-7}$ for the discriminator, both with a weight decay of $1 \times 10^{-6}$. We use the Adam optimizer with $\beta_1 = 0.9$ and $\beta_2 = 0.95$. The training schedule consists of a 1,000 step warm-up, followed by a learning rate decay over 5,001 steps, and a total of 10,000 training steps. The GAN loss weights for both the generator and discriminator are set to $5 \times 10^{-3}$. The training takes around 2 days.
The frequency of different tasks is controlled as follows: the probability ratio for training on MV-oriented mode, input views of 3D-oriented mode, and novel views of 3D-oriented mode tasks is 1:3:1. The ratio for sampling multi-view data versus out-of-distribution data is 2:1.

We use bf16 precision for both training phases. The batch size is 64, using 64 NVIDIA H20 GPUs. For distributed training, we adopt the FSDP (Fully Sharded Data Parallel) strategy and activation checkpointing to improve training efficiency and memory utilization. The prediction of MV-oriented mode is actually $v$-prediction, following the original video diffusion model~\citep{wan2025wan}. We use the flow matching schedule~\citep{lipman2023flow} for both training phases.

\noindent
\textbf{Dataset configuration. }
For both pre-training and post-training, we utilize the following multi-view datasets:
(1) MVImgNet~\citep{yu2023mvimgnet}: an object-centric dataset with a resolution of 480$\times$704;
(2) RealEstate10K~\citep{zhou2018stereo}: an indoor scene dataset with a resolution of 704$\times$480 and frame stride $\in [5,6,7,8,9,10,11,12]$;
(3) DL3DV10K~\citep{ling2024dl3dv}: a general-purpose scene dataset with a resolution of 704$\times$480 and frame stride $\in [2,3,4]$.

For out-of-distribution data during post-training, we employ:
(1) Arbitrary image and text data paired with RealEstate10K and WorldScore camera trajectories: a general dataset with a resolution of 704$\times$480. The images and texts are sampled from a proprietary video dataset.
(2) Echo4O~\citep{ye2025echo} images with WildRGBD~\citep{xia2024rgbd} camera trajectories: a stylized, object-centric dataset with a resolution of 480$\times$704.

\section{Related Works}
\noindent\textbf{Iterative 3D scene generation.}
Recent advances in diffusion models~\citep{rombach2022high,podell2023sdxl,zhang2023adding} have enabled iterative generation of 3D scenes. DiffDreamer~\citep{cai2023diffdreamer} improves multi-view consistency by conditioning on both past and future frames. SceneScape~\citep{fridman2023scenescape}, Text2Room~\citep{hollein2023text2room}, and RGBD2~\citep{lei2023rgbd2} refine mesh-based representations through depth-conditioned diffusion. WonderJourney~\citep{yu2024wonderjourney} leverages point clouds with VLM-guided re-generation. Text2NeRF~\citep{zhang2024text2nerf} and 3D-SceneDreamer~\citep{zhang20243d} address error accumulation by utilizing NeRF~\citep{mildenhall2021nerf} representations. LucidDreamer~\citep{chung2023luciddreamer}, WonderWorld~\citep{yu2025wonderworld}, RealmDreamer~\citep{shriram2025realmdreamer}, and WonderTurbo~\citep{ni2025wonderturbo} accelerate generation and enhance fidelity using 3DGS~\citep{kerbl3Dgaussians}.
While iterative generation methods have made significant progress, they often suffer from cross-view semantic inconsistency. In contrast, data-driven approaches leverage rich cross-view priors to better maintain semantic coherence.

\noindent\textbf{Multi-view-oriented 3D scene generation.}
A major class of data-driven methods adopts a two-stage pipeline: generate multi-view images first, then reconstruct.
CAT3D~\citep{gao2024cat3d} synthesizes novel views via multi-view diffusion, followed by 3D reconstruction. DimensionX~\citep{sun2024dimensionx} generates temporally coherent videos, expands viewpoints through video diffusion, and reconstructs 3D scenes from frames. ODIN~\citep{wallingford2024image} produces trajectory-conditioned novel views for subsequent reconstruction. GenXD~\citep{zhaogenxd} decouples multi-view and temporal features to jointly generate static and dynamic scenes. Bolt3D~\citep{szymanowicz2025bolt3d} outputs colored 3D Gaussians from images and point maps generated by multi-view diffusion. 
Prometheus~\citep{yang2025prometheus} leverages the training paradigm of RGBD latent diffusion models. SplatFlow~\citep{go2025splatflow} jointly learns camera poses and multi-view image distributions from text. Wonderland~\citep{liang2025wonderland} generates continues multi-view latents via video diffusion, then reconstructs scenes using latent-based reconstruction models.

\noindent\textbf{3D-oriented 3D scene generation.}
Another line of work adopts a 3D-oriented pipeline, employing rendering during denoising steps.
DMV3D~\citep{xu2023dmv3d} introduces a large reconstruction-based denoising model based on a triplane NeRF representation, performing denoising through NeRF-based reconstruction and rendering.
Dual3D~\citep{li2024dual3d} proposes a dual-mode multi-view latent diffusion model based on pre-trained image diffusion models and neural surface rendering to reduce training and rendering costs.
Director3D~\citep{li2024director3d} synthesizes pixel-aligned 3D Gaussians directly from latent space using trajectory-conditioned multi-view diffusion, followed by SDS++ refinement.
DiffusionGS~\citep{cai2024baking} presents a diffusion model that outputs pixel-aligned 3DGS at each timestep to ensure 3D consistency.
Cycle3D~\citep{tang2025cycle3d} proposes a unified generation-reconstruction framework, where the 3D reconstruction module is integrated into the multi-step denoising process to further guarantee 3D consistency.

\noindent\textbf{Distillation for diffusion models.} Distillation techniques for diffusion models focus on transferring knowledge from a pretrained teacher model to a more compact and efficient student model. 
Denoising Student~\citep{luhman2021knowledge} achieves this by training a single-step generator to minimize the RMSE between the outputs of the teacher and student models.
Consistency Model~\citep{song2023consistency} enables trajectory distillation, allowing the student to mimic the teacher's denoising process across multiple steps.
Adversarial Diffusion Distillation (ADD)~\citep{sauer2024adversarial}, Latent Adversarial Diffusion Distillation (LADD)~\citep{sauer2024fast}, Adversarial Post-Training (APT)~\citep{lin2025diffusion}, and Autoregressive Adversarial Post-Training (AAPT)~\citep{lin2025autoregressive} further enhance distillation by introducing adversarial objectives to improve the student performance.
Distribution Matching Distillation (DMD)~\citep{yin2024one} formulates the distillation objective as optimizing the reverse KL-divergence between the student and teacher distributions.
DMD2~\citep{yin2024improved} extends this framework by incorporating a GAN-based objective and supporting for multi-step generators, further improving the flexibility and effectiveness of the distillation process.

\section{Limitations}

While the proposed \modelname~demonstrates strong capabilities in generating high-fidelity and efficient 3D scenes, several limitations remain.
First, despite increasing the number of views, the diversity and scale of generated scenes are still constrained by the coverage of existing datasets.
Second, the model currently struggles with accurately generating fine-grained geometry, mirror reflections, and articulated objects.
These issues may be alleviated by incorporating depth priors~\citep{yang2024depth,yang2024depthv2,chen2025video} and more 3D-aware structural information~\citep{jiang2025anysplat,wang2025splatvoxel} to further enhance the quality of our pixel-aligned 3D Gaussians.

\section{RGBD Rendering Results}

\begin{figure}[t]
    \centering
    \includegraphics[width=1\linewidth]{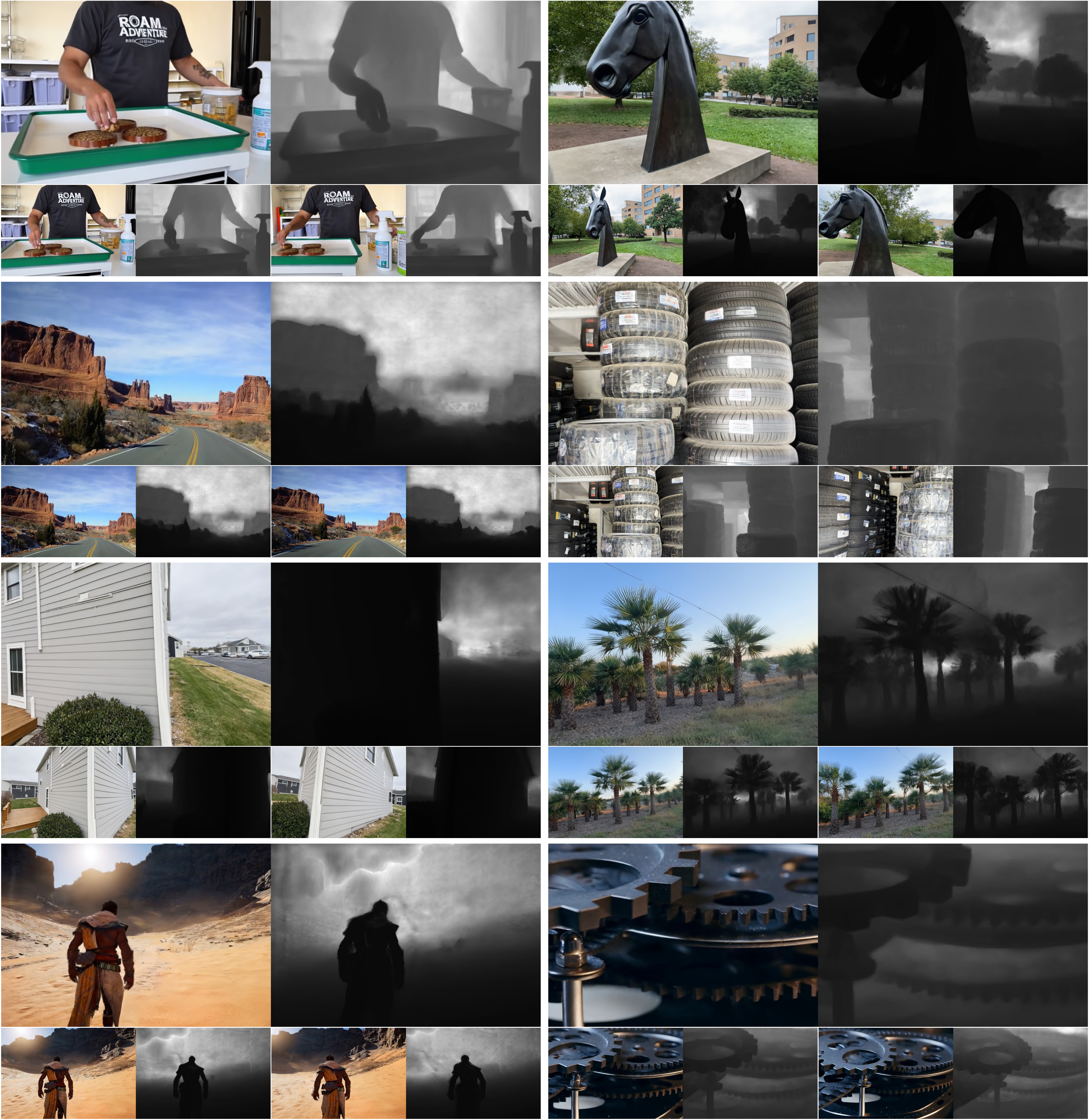}
    \caption{{\bf RGBD rendering results.}}
    \label{fig:depth_results}
\end{figure}

While \modelname~does not explicitly incorporate depth supervision, the 3DGS outputs inherently enable the export of depth maps. 
In this regard, we present several RGBD rendering results in Fig.~\ref{fig:depth_results}. 
This serves to demonstrate that our model is capable of learning meaningful depth geometric information solely via image supervision.

\section{More Results}

\begin{figure}[t]
    \centering
    \includegraphics[width=0.96\linewidth]{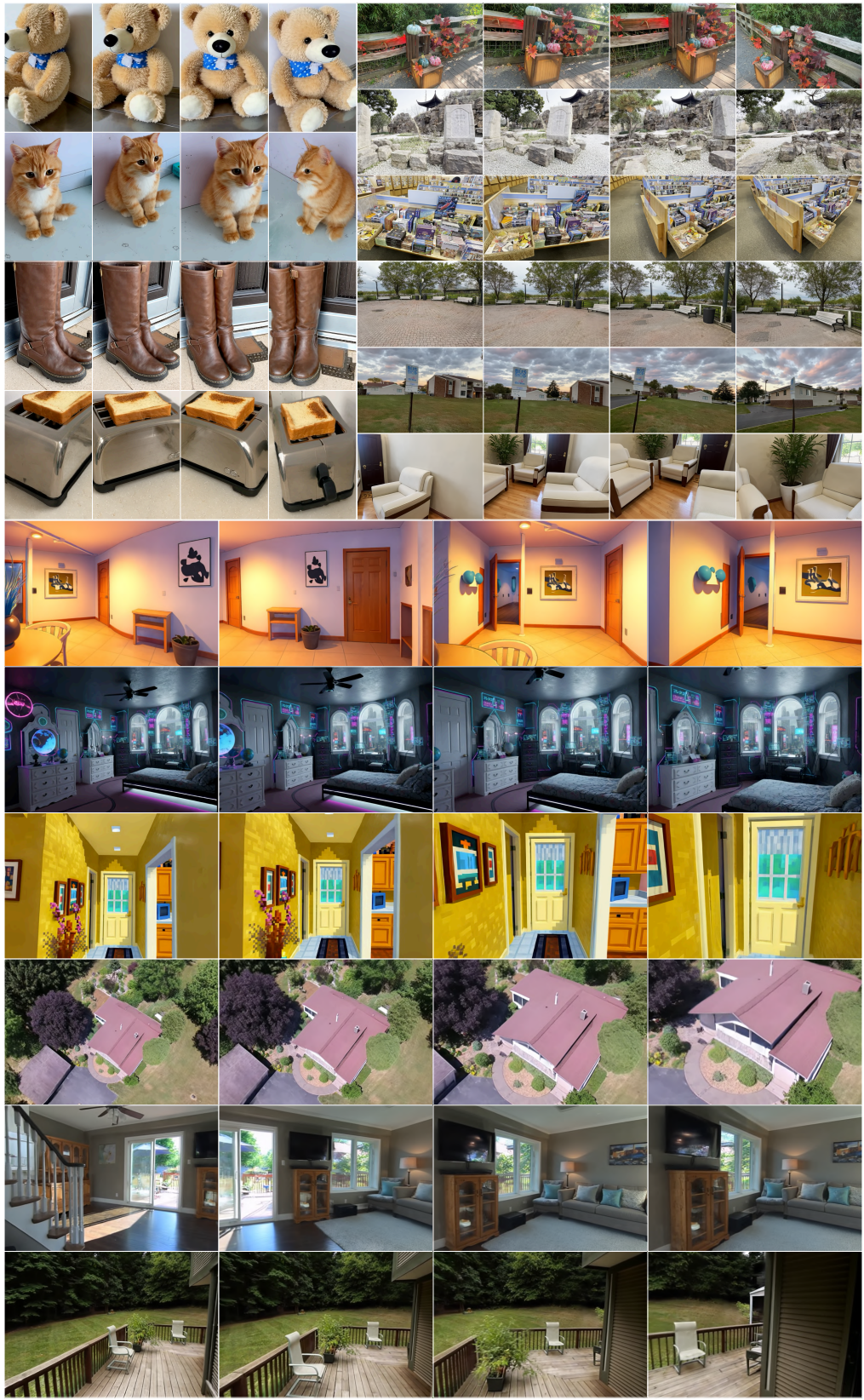}
    \caption{{\bf More generation results.} All images are rendered with generated 3DGS.}
    \label{fig:more_results}
\end{figure}

We provide more generation results in Fig.~\ref{fig:more_results}, including object-centric, indoor, outdoor, realistic, and stylized scenes, to demonstrate the strong and generalizable generation ability of our model. 

For video rendering results, please kindly refer to our project page.

\end{document}